\documentclass[10pt,twocolumn,letterpaper]{article}

\usepackage[pagenumbers]{cvpr}           %

\usepackage[dvipsnames]{xcolor}

\newcommand{\br}{\mathbf{r}}

\newcommand{\figref}[1]{Fig.~\ref{#1}}
\newcommand{\secref}[1]{Sec.~\ref{#1}}

\newcommand{\eqnref}[1]{Eq.~\eqref{#1}}
\newcommand{\tabref}[1]{Table~\ref{#1}}

\makeatletter
\DeclareRobustCommand\onedot{\futurelet\@let@token\@onedot}
\def\@onedot{\ifx\@let@token.\else.\null\fi\xspace}
 
\def\ie{i.e\onedot} 
 
\def\etc{etc\onedot}

\def\etal{et~al\onedot}

\makeatother

\newcommand{\boldparagraph}[1]{\vspace{0.3em}\noindent{\bf #1.}}

\renewcommand{\paragraph}[1]{\boldparagraph{#1}}

\definecolor{darkgreen}{rgb}{0,0.7,0}
\definecolor{newyellow}{rgb}{1,0.8,0.05}
\definecolor{newgreen}{rgb}{0.2,0.8,0.2}
\definecolor{Gray}{gray}{0.85}

\makeatletter
\def\adl@drawiv#1#2#3{%
        \hskip.5\tabcolsep
        \xleaders#3{#2.5\@tempdimb #1{1}#2.5\@tempdimb}%
                #2\z@ plus1fil minus1fil\relax
        \hskip.5\tabcolsep}
\newcommand{\cdashlinelr}[1]{%
  \noalign{\vskip\aboverulesep
           \global\let\@dashdrawstore\adl@draw
           \global\let\adl@draw\adl@drawiv}
  \cdashline{#1}
  \noalign{\global\let\adl@draw\@dashdrawstore
           \vskip\belowrulesep}}
\makeatother

\definecolor{cvprblue}{rgb}{0.21,0.49,0.74}
\usepackage[pagebackref,breaklinks,colorlinks,citecolor=cvprblue]{hyperref}
\usepackage{algorithm}
\usepackage[noend]{algpseudocode}
\usepackage[accsupp]{axessibility} %
\usepackage{multicol}
\usepackage{multirow}

\def\confName{CVPR}
\def\confYear{2024}

\title{3D Neural Edge Reconstruction}

\author{Lei Li{$^{1}$}~~~~~ Songyou Peng{$^{1,2\dagger}$}~~~~~ Zehao Yu{$^{3,4}$}~~~~~ Shaohui Liu{$^{1}$}~~~~~Rémi Pautrat{$^{1, 6}$}\\Xiaochuan Yin{$^{5}$}~~~~~ Marc Pollefeys{$^{1, 6}$}\\\vspace{-13pt}{\small~}\\
{$^{1}$}ETH Zurich~~~~~{$^{2}$}MPI for Intelligent Systems, Tübingen~~~~~{$^{3}$}University of Tübingen~~~~~\\{$^{4}$}Tübingen AI Center~~~~~~~~~~~~~{$^{5}$}Utopilot~~~~~~~~~~~~~{$^{6}$}Microsoft\\
{\href{https://neural-edge-map.github.io}{neural-edge-map.github.io}}
}

\begin{document}
\maketitle
\let\thefootnote\relax\footnotetext{$^{\dagger}$ Corresponding author}
\begin{abstract}
Real-world objects and environments are predominantly composed of edge features, including straight lines and curves. Such edges are crucial elements for various applications, such as CAD modeling, surface meshing, lane mapping, \etc. However, existing traditional methods only prioritize lines over curves for simplicity in geometric modeling. To this end, we introduce EMAP, a new method for learning 3D edge representations with a focus on both lines and curves. Our method implicitly encodes 3D edge distance and direction in Unsigned Distance Functions (UDF) from multi-view edge maps. On top of this neural representation, we propose an edge extraction algorithm that robustly abstracts parametric 3D edges from the inferred edge points and their directions. Comprehensive evaluations demonstrate that our method achieves better 3D edge reconstruction on multiple challenging datasets. We further show that our learned UDF field enhances neural surface reconstruction by capturing more details.

\end{abstract}

\section{Introduction}

\begin{center}
\emph{The straight line belongs to men, the curved one to God.
}\\
\hfill--- Antonio Gaudí
\vspace{-0.5em}
\end{center}
This sentiment is evident in the visual composition of our environments. While straight lines are common in man-made scenes such as walls, windows, and doors~\cite{liu20233d}, curves are more general and ubiquitous from cups, bridges, architectures, to Gothic arts.
\emph{Edges}, which are composed of both lines and curves, are the fundamental elements of visual perception. 
Therefore, accurate edge modeling is crucial for understanding the geometry and structure of our 3D world.

Conventional approaches on 3D reconstruction typically involve inferring dense geometry and abstracting meshes from 2D images~\cite{yariv2020multiview,yariv2021volume,Oechsle2021ICCV,Schonberger2016ECCV,Goesele2007ICCV,Seitz2006CVPR}. 
However, the presence of 3D edges offers substantial advantages. 
First, edges are naturally compact representations that capture the salient features oftentimes around geometric boundaries, which are good indicators for more lightweight and adaptive meshing and 3D modeling with comparably less redundancy.  
Secondly, in contrast to dense surface modeling from images, 3D edges are unaffected to illumination changes, thus exhibiting better reliability on multi-view reconstruction.
Last but not least, 3D edges serve as a universal representation in real-world scenarios, and can be potentially integrated into many applications such as lane mapping~\cite{qin2021light, qiao2023online, cheng2021road, wang2023road, li2023topology}, motion forecasting~\cite{fang2020tpnet, ettinger2021large, shi2022motion}, medical imaging~\cite{rorden2022improving}, \etc.

\begin{figure}[t]
    \centering
    \subfloat[An Indoor Scene]{%
        \includegraphics[width=0.48\linewidth]{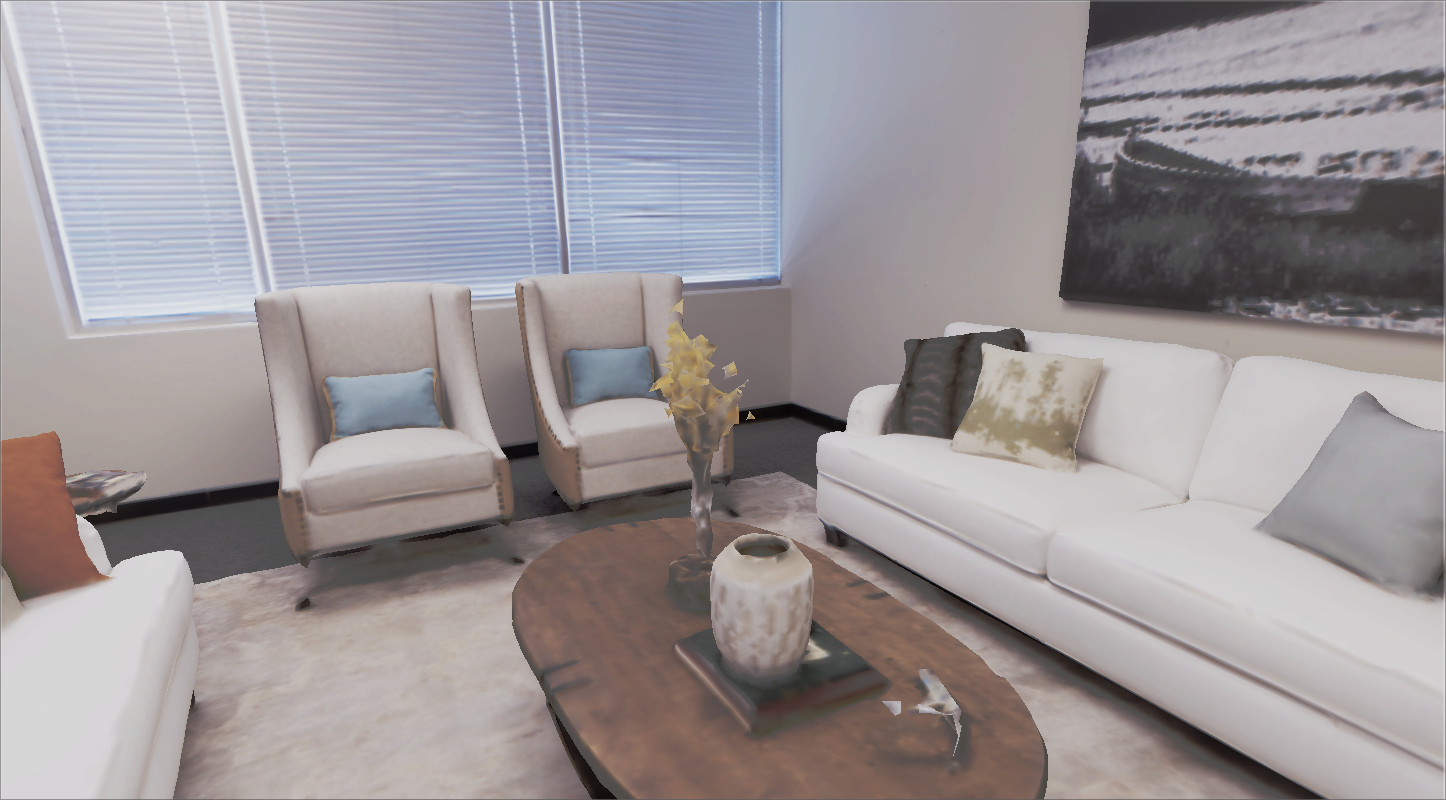}
        \label{fig:teaser_image}
    }
    \subfloat[LIMAP~\cite{liu20233d}]{%
        \includegraphics[width=0.48\linewidth]{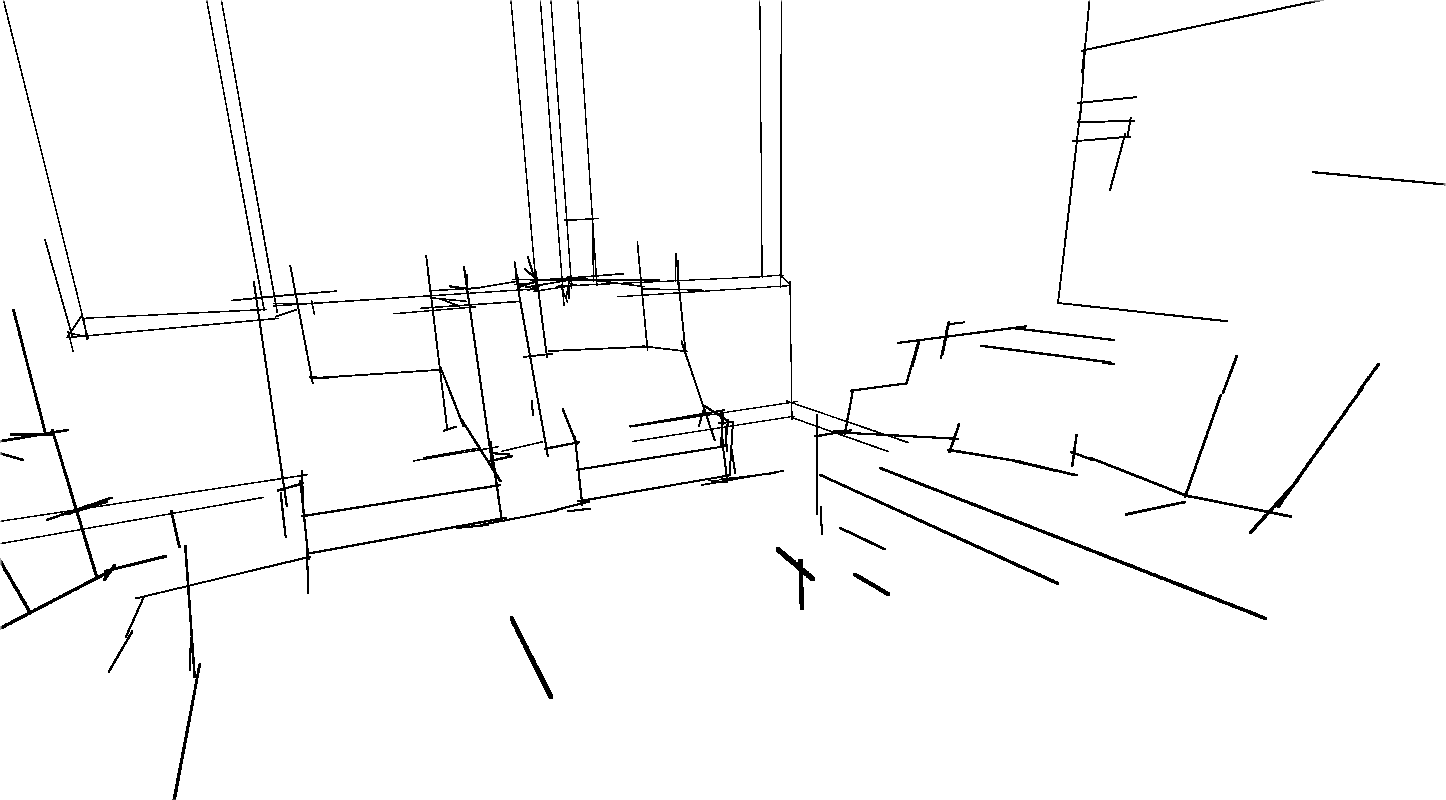}
        \label{fig:teaser_limap}
    }\\
    \subfloat[NEAT~\cite{xue2023volumetric}]{%
        \includegraphics[width=0.48\linewidth]{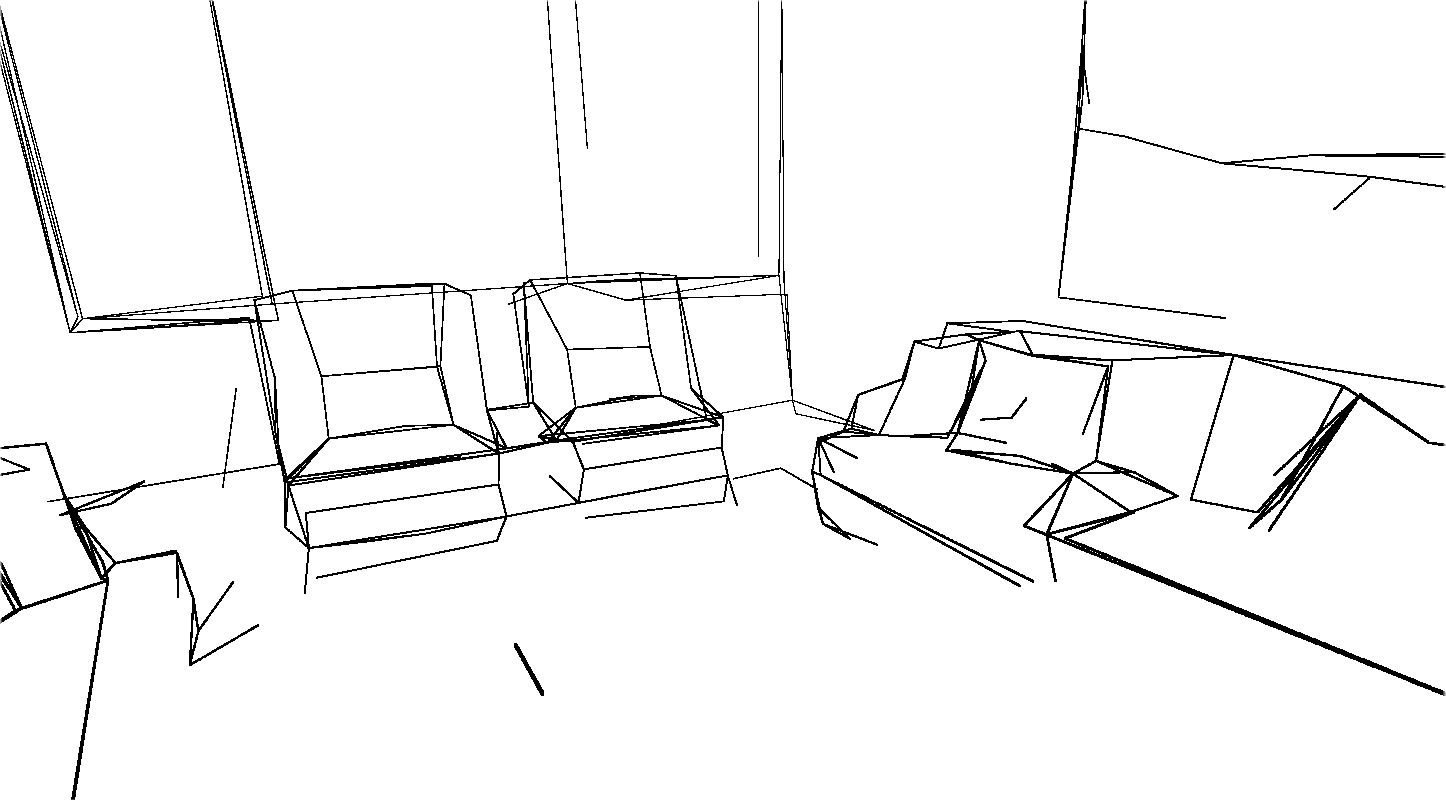}
        \label{fig:teaser_neat}
    }
    \subfloat[EMAP (\textbf{Ours})]{%
        \includegraphics[width=0.48\linewidth]{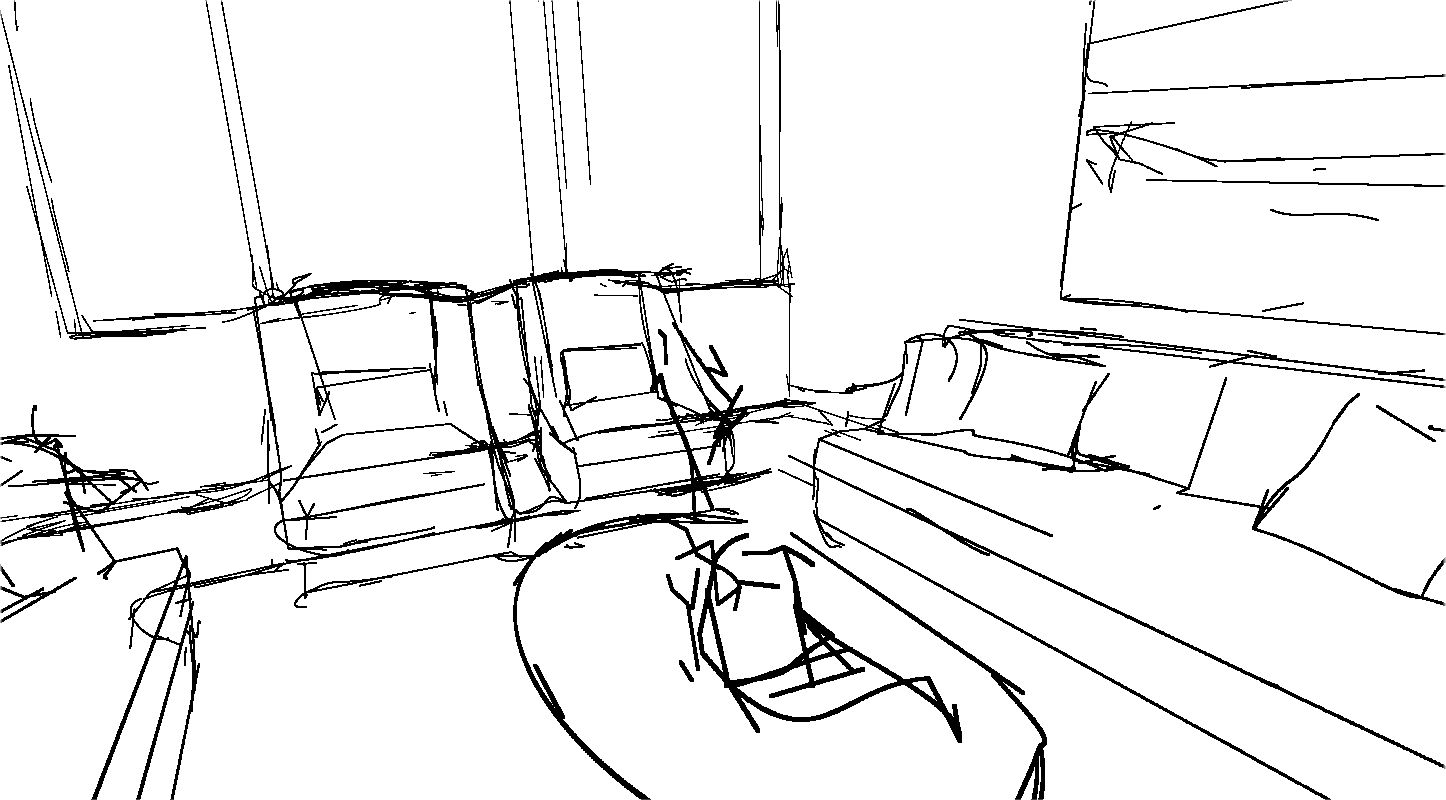}
        \label{fig:teaser_ours}
    }
    \caption{\textbf{Example 3D edge reconstruction on Replica~\cite{straub2019replica}.} While prior methods such as LIMAP~\cite{liu20233d} and NEAT~\cite{xue2023volumetric} only reconstruct distinctive line segments, our method generates a more complete 3D edge map combining both line and curve features.
    }
    \label{fig:teaser}
    \vspace{-1.5em}
\end{figure}

The reconstruction of 3D edges is conventionally approached by matching their 2D observations across views.
While Bignoli \etal~\cite{bignoli2018edgegraph3d} proposed edge point matching using the sparse map from Structure-from-Motion (SfM), it is inherently ill-posed due to its heavy reliance on cross-view edge correspondences, which are generally sparse and prone to ambiguity.
Recent works have also improved the quality of 3D line reconstruction~\cite{hofer2014improving,wei2022elsr,liu20233d,xue2023volumetric}, but primarily excel in specific scenes where straight lines dominate. While general real-world environments with curved structures pose more challenges, recent progress on 2D detection and matching is mostly limited to point and line features and thus inapplicable to such scenarios.

A recent work NEF~\cite{ye2023nef} made a significant step forward in learning 3D curves from multi-view 2D edge observations. Inspired by the recent success of neural radiance field (NeRF)~\cite{mildenhall2020nerf}, they introduce a neural edge density field and show decent results in reconstructing edges for simple objects.
Nevertheless, their proposed edge density field has an inherent bias in edge rendering, leading to less accurate reconstruction.
Moreover, its fitting-based edge parameterization process not only requires tedious tuning to specific data, but also struggles with its scalability to larger and more complex scenes. 
This motivates us to develop a more robust system for 3D edge mapping from 2D observations, which would benefit a wide range of downstream tasks.

Towards this goal, we introduce \emph{EMAP}, a novel approach for accurate 3D edge reconstruction from only 2D edge maps.
EMAP comprises the following steps.
Firstly, we learn the neural unsigned distance function (UDF) to implicitly model 3D edges, utilizing an unbiased rendering equation to mitigate the inaccuracies observed in NEF.
Secondly, once learned, we can obtain the unsigned distance and normal for each point in the space, so a set of precise edge points with directions can be extracted. 
Finally, based on the guidance of every edge point's location and direction, we design a simple yet robust algorithm for parametric line and curve extraction, that can be applied across various challenging scenarios.
Our comprehensive evaluations of EMAP, from synthetic CAD models to real-world indoor and outdoor scenes, show its superior performance in 3D edge reconstruction.
In addition, we also observe that initializing the optimization process of the recent neural implicit surface reconstruction method with our trained UDF field enables the reconstructing of better details.

Overall, the contributions of this paper are as follows:
\begin{itemize}
    \item We propose EMAP, a 3D neural edge reconstruction pipeline that can learn accurate 3D edge locations and directions implicitly from multi-view edge maps.
    \item We develop a 3D edge extraction algorithm to robustly connect edge points with edge direction guidance. 
    \item We show that our model can generate complete 3D edge maps and help optimize dense surfaces.
\end{itemize}

\section{Related Work}

\boldparagraph{Geometry-based 3D Line Reconstruction}
As a pioneering work, Bartoli and Sturm \cite{bartoli2005structure} introduces a full SfM system using line segments, which is later improved under manhattan assumption \cite{schindler2006line} and in stereo systems \cite{chandraker2009moving}. Recently, with the developments of line detections \cite{pautrat2021sold2,xue2023holistically,pautrat2023deeplsd} and matching \cite{pautrat2021sold2,abdellali2021l2d2,pautrat_suarez_2023_gluestick} thanks to the advent to deep learning, several works have attempted to revisit the line mapping problem through graph clustering \cite{hofer2017efficient}, leveraging planar information \cite{wei2022elsr} and incorporating into SLAM systems \cite{marzorati2007integration,zuo2017robust,he2018pl,lim2022uv,shu2023structure}. In particular, recent work LIMAP \cite{liu20233d} introduces a robust 3D line mapping system with structural priors which can adapt to different existing line detectors and matchers. Despite these advances, all the works are limited to straight lines and often produce segmented small lines when it comes to curves. In contrast, edges are generally easier to detect and are redundantly present in most scenes. In this project, rather than relying on lines, we build our 3D mapping system using robust 2D edge maps.

\boldparagraph{Learning-based 3D Line/Curve Reconstruction} 
In contrast to geometry-based methods, 
some approaches~\cite{wang2020pie,liu2021pc2wf, zhu2023nerve} shifted their focus to directly extract parametric curves from given edge point clouds. Typically, they require key-point detection, clustering, and linkage. 
Even under the relaxed setting, it is still challenging to generate clean parametric curves due to the complex connectivity of curves and imperfect point clouds~\cite{ye2023nef}.
To address this limitation, NEF~\cite{ye2023nef} integrates NeRF~\cite{mildenhall2020nerf} for edge mapping from multi-view images, extracting 3D curves from the learned neural edge field through a carefully designed post-processing. While NEF achieves decent performance on CAD models, it is constrained to simple and low-precision object-level edge mapping. 
A concurrent work, NEAT~\cite{xue2023volumetric}, utilizes VolSDF~\cite{yariv2021volume} to build dense surfaces and incorporates a global junction perceiving module to optimize 3D line junctions with 2D wireframe supervision. Although NEAT can produce 3D wireframes, it is restricted to modeling line segments only. Additionally, their need for textured objects is a limitation. 
By contrast, we use the unisgned distance function (UDF) to represent edges, enabling the construction of both line segments and curves without the necessity for target textures.
We further show that our method can faithfully reconstruct edges for complex scenes.

\boldparagraph{Neural Implicit Representations}
\label{related_representation} 
Neural implicit representations have emerged as a powerful tool for a spectrum of computer vision tasks, including object geometry representation~\cite{Mescheder2019CVPR,Park2019CVPR,Chen2019CVPR,Xu2019NIPS,Yariv2020NEURIPS,Oechsle2021ICCV,Wang2021NEURIPS,Yariv2021NEURIPS,Saito2019ICCV,Liu2020CVPR,Peng2021NEURIPS}, scene reconstruction~\cite{Peng2020ECCV,Jiang2020CVPR,Chabra2020ECCV,Zhu2022CVPR,Zhu2024THREEDV,yu2022monosdf,Yu2022SDFStudio}, novel view synthesis~\cite{mildenhall2020nerf,Zhang2020ARXIV,Martin2021CVPR,Reiser2021ICCV} and generative modelling~\cite{Schwarz2020NEURIPS,Chan2022CVPR,Niemeyer2021CVPR}. 
Recent works~\cite{wang2021neus, yariv2021volume, yu2022monosdf, wang2022hf, long2023neuraludf} show impressive high-fidelity reconstruction by learning the implicit signed distance function (SDF). However, the SDF representation constrains to modeling closed, watertight surfaces. In contrast, NeuralUDF~\cite{azinovic2022neural} exploits UDF to represent surfaces, offering a higher degree of freedom to represent both closed and open surfaces. 
We find UDF as a suitable representation to model edges implicitly, in comparison to SDF used in NEAT~\cite{xue2023volumetric} and edge volume density from NEF~\cite{ye2023nef}.

\begin{figure}[ht]
  \begin{center}
  \includegraphics[width=1.0\linewidth]{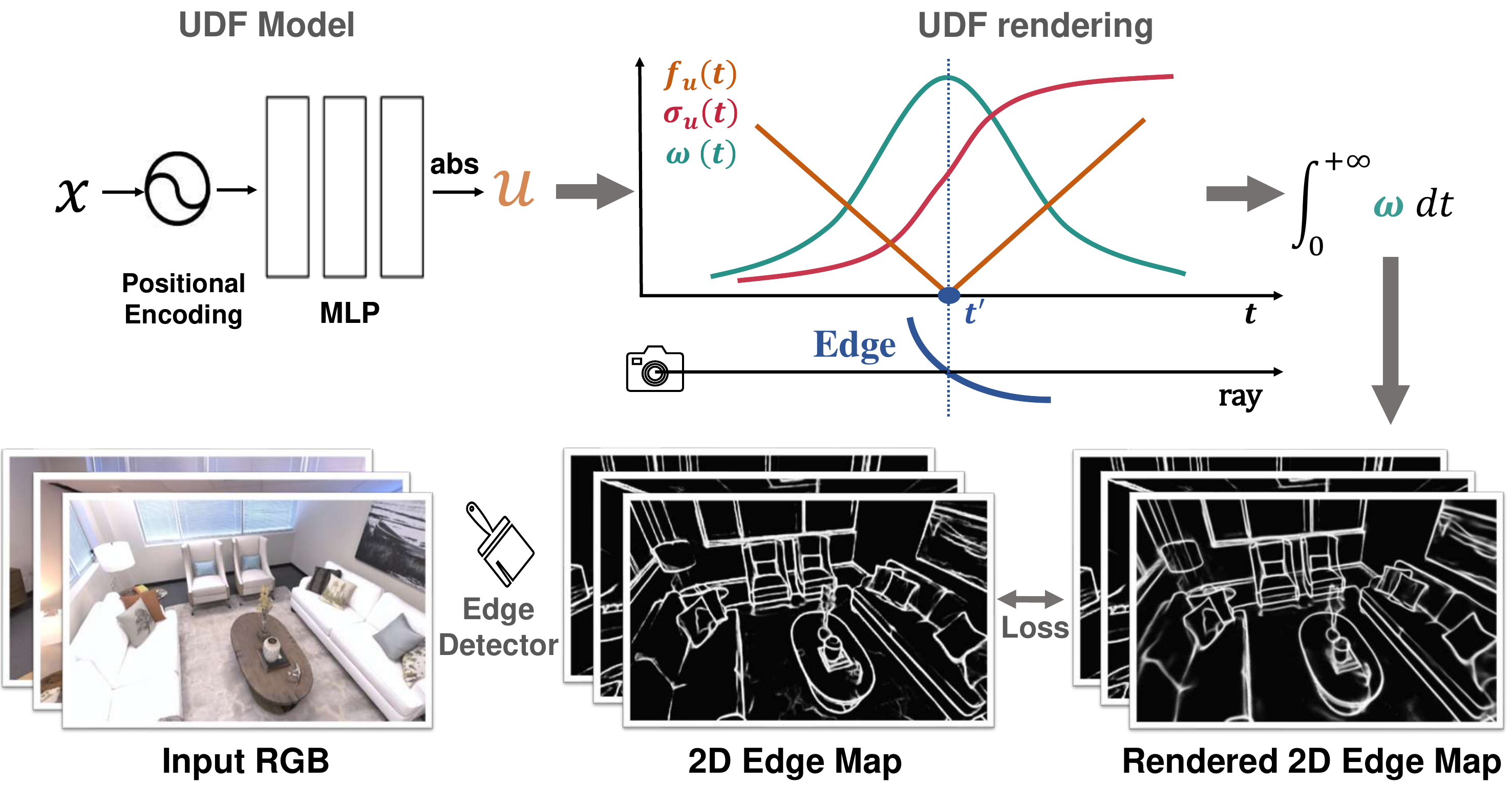}  
  \end{center}

  \vspace{-2.0em}
  \caption{\textbf{UDF learning overview.} We utilize a vanilla NeRF~\cite{mildenhall2020nerf} MLP that outputs absolute values to model the 3D UDF field. Edge maps are rendered using a density-based edge neural rendering technique, combined with an unbiased UDF rendering approach to eliminate bias. Our primary supervision comes from 2D edge maps predicted by a pre-trained edge detector.}
  \label{fig:UDF_pipeline}
  \vspace{-1.0em}
\end{figure}

\section{Method}
Our goal is to build a 3D edge map from multi-view posed 2D edge maps. To this end, we first introduce our edge representation and edge field learning in~\secref{sec:udf_detail}. Next, we present our 3D parametric edge extraction from the learned edge representations in~\secref{sec:edge_extraction}.

\subsection{Edge Field with Unsigned Distance Functions}\label{sec:udf_detail}
\boldparagraph{Multi-view Edge Maps} Since edge maps are generally invariant to illumination changes and are more robustly detected across various scenes than lines,
our method utilizes multiple posed 2D edge maps as inputs. We apply pre-trained edge detectors to predict an edge map $E$ for each input RGB image. 
Each pixel of $E$ has a value within $[0,1]$, indicating its probability of being an edge.

\boldparagraph{Density-based Edge Neural Rendering} 
We use an unsigned distance function (UDF) to represent edges, denoted as $f_u$. This function computes the unsigned distance from a given 3D point to the nearest edge.
The UDF is defined as:
\begin{equation}
    f_u: \mathbb{R}^3 \rightarrow \mathbb{R} \quad \mathbf{x} \mapsto u=\operatorname{UDF}(\mathbf{x}) \,,
\end{equation}
where x is a 3D point and $u$ is the corresponding UDF value.

To render an edge pixel in a certain view, we trace a camera ray $\mathbf{r}(t)=\mathbf{o}+t \mathbf{d}$. This ray originates from the camera's center $\mathbf{o}$ and extends in direction $\mathbf{d}$~\cite{mildenhall2020nerf}.
To apply volume rendering for edge modeling, it is necessary to establish a mapping $\Omega_u$~\cite{wang2021neus, long2023neuraludf} that transforms the distance function $f_u(\br(t))$ into volume density $\sigma_u(t)$ as 
\begin{equation}
\sigma_u(t) = \Omega_u(f_u(\br(t)))\,.
\end{equation}
In the rendering equation, the transmittance $T(t)$ and weight $\omega(t)$ along the camera ray $\mathbf{r}$ are accumulated as
\begin{equation}
    T(t) =\exp \left(-\int_0^t \sigma_u(v) d v \right),\;\;\; w(t) =T(t) \cdot \sigma_u(t)\,.
\label{eq:transmittance}
\end{equation}
To effectively handle appearance changes under different viewing angles, most neural field-based surface reconstruction~\cite{Oechsle2021ICCV, yariv2020multiview, wang2021neus,Yu2022SDFStudio} disentangles geometry and appearance. In contrast, edge maps are generally unaffected by lighting, making them view-independent.
Therefore, this simplifies the rendering process for edge maps. as it only requires the accumulation of view-independent, density-based weights $w$ along a ray $\br$.
Now, the rendered edge value $\hat{E}$ along ray $\mathbf{r}$ is formulated as:
\begin{equation}
\hat{E}(\mathbf{r}) = \int_0^{+\infty} w(t) d t =1-T(+\infty)\, ,
\label{eq:rendering}
\end{equation}
\eqnref{eq:rendering} establishes the connection between rendered edge values and the transmittance at the end of the camera rays. Intuitively, this means that the rendered edge value is $1$ when the camera ray hits an edge in 3D space, and 0 otherwise. Please refer to the supplements for more details.

\boldparagraph{Unbiased Density Functions for UDF Rendering}\label{sec:unbiased}
NEF~\cite{ye2023nef} also uses volume rendering for rendering edges. Unlike ours, they utilize edge density to represent edges and an additional network to predict edge values. However, this approach introduces an inherent bias in edge rendering. 
Similar to the naive solution presented in NeuS~\cite{wang2021neus}, the issue comes from the weight function $w$ in~\eqnref{eq:transmittance}, where its local maximum does not coincide with the actual intersection point of the camera ray and the edges.

To address this issue, we incorporate unbiased UDF rendering~\cite{long2023neuraludf} into our density-based edge rendering framework. As proved in NeuS, density function $\sigma_u$ should increase monotonically to make the weight function unbiased. However, UDF values are not monotonous along a ray~\cite{long2023neuraludf}. To adapt the unbiased density function $\Omega_s$, which is originally induced in NeuS~\cite{wang2021neus}, for UDF use, the monotonically increased density function $\sigma_u$~\cite{long2023neuraludf} is formulated as
\begin{equation}
\sigma_u(t) =\Psi(t)\cdot\Omega_s\left(f_u(\mathbf{r}(t))\right)+(1-\Psi(t)) \cdot \Omega_s\left(-f_u(\mathbf{r}(t))\right) \,,
\end{equation}
where $\Psi(t)$ is a differentiable visibility function designed in \cite{long2023neuraludf} to capture the monotonicity change in UDF. $\Psi$ is 0 behind the intersection point between the camera ray and the hit edge, and is 1 before the intersection point. Besides, $\Psi(t)$ is differentiable around the intersection point to make the UDF optimization more stable.

\boldparagraph{Ray Sampling Strategy}
A key characteristic of 2D edge maps is their significant sparsity, with edges occupying a much smaller area compared to non-edge regions.
To enhance training efficiency and stability, we apply an importance sampling strategy for camera rays, with 50\% of rays uniformly sampled from edge areas in the edge maps and the remaining 50\% from non-edge areas.
Such a sampling strategy ensures that our UDF field training is concentrated on edge areas, thereby substantially speeding up the training process.
Additionally, our sampling strategy offers an elegant solution to the issue of occlusion, a challenge noted in~\cite{ye2023nef}.
The rendered edge maps might contain edges not present in the input edge images due to occlusion.
In contrast to the complicated occlusion handling strategy introduced in~\cite{ye2023nef},
our approach inherently alleviates this challenge 
by focusing the training on points from the visible edges presented in the input edge maps.

\boldparagraph{Loss Functions} The total loss function can be written as:
\begin{equation}
\mathcal{L}_{\text {total}}=\mathcal{L}_{\text {edge}} + \lambda\mathcal{L}_{\text {eik}} \,, \label{eq:total_loss}
\end{equation}
where $\mathcal{L}_{\text{edge}}$ represents the Mean Square Error (MSE) between the rendered and input edge images. $\mathcal{L}_{\text {eik}}$ denotes the Eikonal loss~\cite{gropp2020ICML}, which promotes the learned UDF to be physical distance.
$\lambda$ is used to balance these losses.

\subsection{3D Parametric Edge Extraction}\label{sec:edge_extraction}
\begin{figure}[!t]
  \begin{center}
  \includegraphics[width=0.85\linewidth]{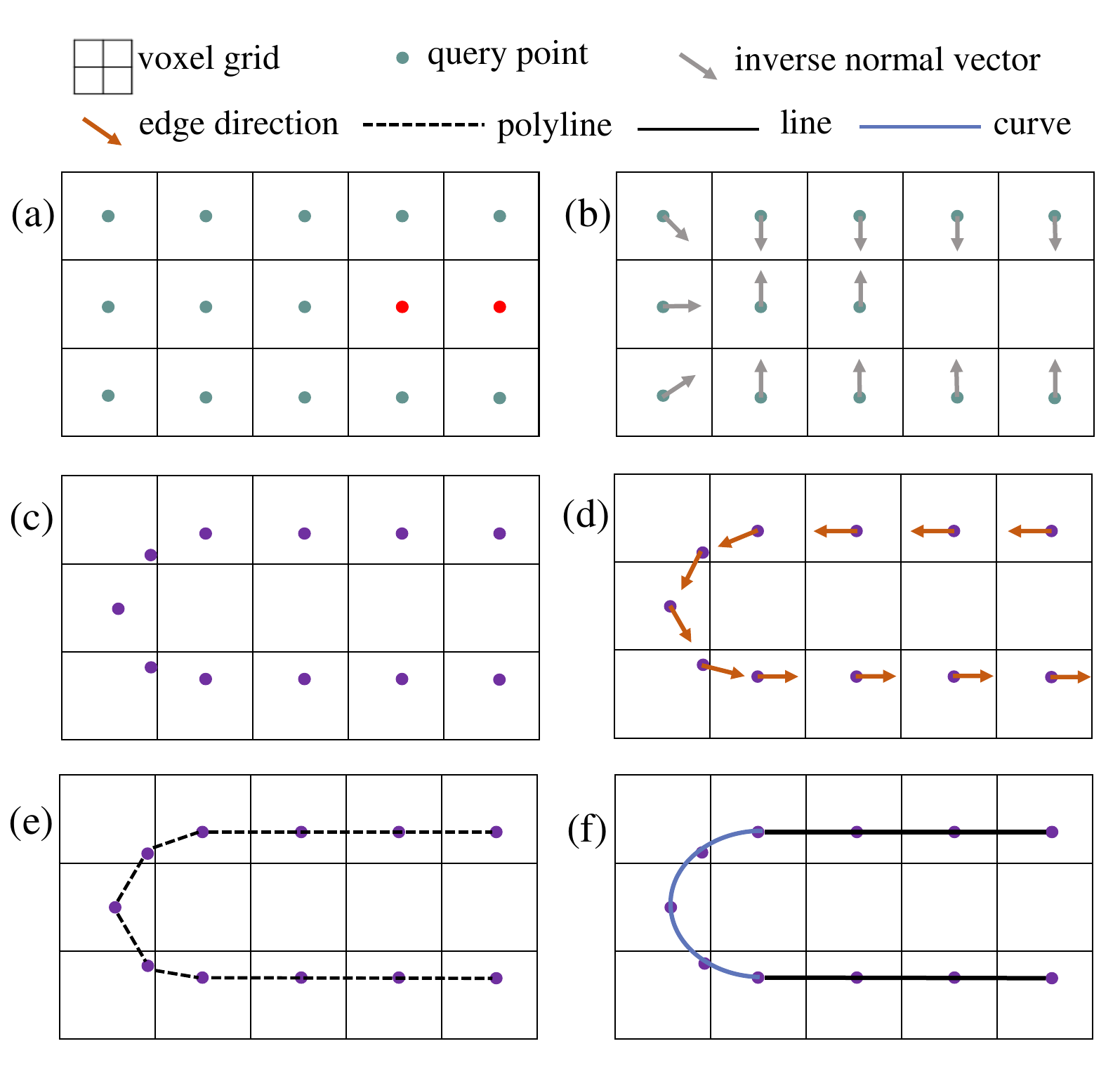}  
  \end{center}
  \vspace{-1.5em}
  \caption{\textbf{Illustration of our 3D parametric edge extraction steps.} 
  For simplify, our schematic is depicted in the 2D plane.
  Our 3D edge extraction algorithm comprises five main stages: point initialization (a), point shifting (b to c), edge direction extraction (c to d), point connection (d to e), and edge fitting (e to f).
  }
  \label{fig:postprocessing}
  \vspace{-1.0em}
\end{figure}

With UDF learning, edge locations are implicitly encoded within the UDF field. However, accurately extracting edge points from the UDF field is non-trivial due to the absence of a real zero-level set in the UDF field. Additionally, formulating these edge points into parametric edges poses significant challenges due to the complex connections of edges. 
To extract points from the learned density field, NEF~\cite{ye2023nef} selects points with edge density values greater than a specified threshold, $\epsilon$. This approach leads to an approximated edge point set that is $\epsilon$-bounded~\cite{long2023neuraludf}. While this method effectively generates comprehensive point clouds, the $\epsilon$-bounded point set does not align accurately with the actual edge locations.

\begin{figure}[!t]
  \begin{center}
  \includegraphics[width=1.0\linewidth]{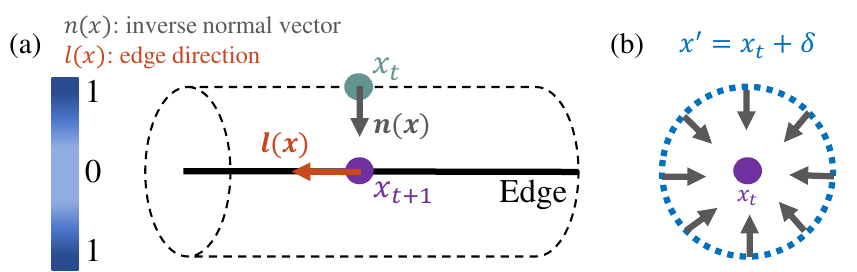}  
  \end{center}
  \vspace{-1.0em}
  \caption{\textbf{Illustration of the overview (a) and the cross-section (b) of UDF field.} (a) In UDF field, edge points are ideally located at the zero-level set, with UDF values being larger away from these points. A query point $x_t$ can be precisely shifted to a more accurate position $x_{t+1}$ by following the UDF value and the inverse normal vector $n(x)$. 
  The edge direction $l(x)$ aligns with the tangent direction at the edge point $x_{t+1}$. 
  (b) The inverse normal vectors of all surrounding points on the cross section are pointing towards the query point.}
  \label{fig:optimizing_UDF}
  \vspace{-1.0em}
\end{figure}
\label{point_shifting}
To eliminate the error in edge point extraction, we leverage the physical property of UDF that reflects real-world distances to the edges. Specifically, we develop a 3D edge extraction algorithm composed of five main stages: point initialization, point shifting, edge direction extraction, point connection, and edge fitting, as illustrated in \figref{fig:postprocessing}. This algorithm takes the trained UDF field as input and outputs parametric 3D edges, including line segments and curves. 

\boldparagraph{Point Initialization} Under eikonal loss supervision, the optimized UDF values represent physical distances to the nearest edges. To initialize potential edge points, we begin with the center points of all voxel grids and obtain their UDF values from the UDF field. Subsequently, we eliminate query points whose UDF values exceed a specified threshold $\epsilon'$ (red points in~\figref{fig:postprocessing}~(a)). 

\boldparagraph{Point Shifting} As illustrated in \figref{fig:optimizing_UDF}~(a), the normalized inverse gradient of the UDF field indicates the inverse normal vector pointing towards edges. Drawing inspiration from OccNet~\cite{Mescheder_2019_CVPR}, we refine the point $x$ iteratively towards the edge using its distance and inverse normal direction:
\begin{equation}
    x_{t+1} \Leftarrow x_t - f_u(x_t) \cdot \frac{\nabla f_u(x_t)}{\left\|\nabla f_u(x_t)\right\|}\,,
\end{equation}
where $t$ denotes the $t$-th iteration. 
As a result of this iterative process, the initial points converge to the edge center (from \figref{fig:postprocessing}~(b) to \figref{fig:postprocessing}~(c)).

\boldparagraph{Edge Direction}
Establishing connections between edge points is a crucial step in constructing parametric edges. While most methods~\cite{ye2023nef, cherenkova2023sepicnet, qiao2023online} estimate parameters through least-squares fitting of lines/curves on extracted points, this fitting-based approach for edge extraction is not always robust or accurate. In contrast, inspired by \cite{xue2019learning, pautrat2023deeplsd}, we find that combining the edge direction field with the edge distance field can robustly produce edge parameters.
Given that inverse normal vectors invariably point towards edges (see \figref{fig:optimizing_UDF}~(b)), we first devise an edge direction extraction method based on this set of inverse normal vectors. Specifically, for a query point $x$, we introduce minor shifts set $\{\delta\}_N$ with size of N to generate an adjoining point set $\{x'\}_N$, where $\{x'\}_N = x + \{\delta\}_N$. The inverse normal vectors of these points, denoted as $\{n\}_N$, are obtained from the learned UDF field.
The edge direction, denoted as $l$, is identified as the null space of $\{n'_i\}$, since the edge direction is perpendicular to all inverse normal vectors in $\{n\}_N$. %
Therefore, $l$ can be extracted with singular value decomposition (SVD):
\begin{equation}
A = U \Sigma V^T\,, \quad
l = V[:, \text{argmin}(\Sigma)]\,,
\end{equation}
where $A$ is the matrix representation of $\{n\}_N$ and $l$ corresponds to the eigenvector associated with the smallest eigenvalue. Note that $N$ should be sufficiently large to ensure the stability of the extracted edge direction. Unlike DeepLSD~\cite{pautrat2023deeplsd}, we can obtain a precise edge direction field without relying on any 2D direction supervision.

\boldparagraph{Point Connection} 
After accurately determining the edge point locations and directions, we proceed to connect these edge points guided by the edge direction to create polylines (\figref{fig:postprocessing}~(d) to (e)).
Specifically, we begin by selecting candidate points and then compute directional errors for points adjacent to these candidates. Based on these directional errors, candidate points are connected to its best-matched neighboring point that growing direction aligns best with its extracted edge direction, \ie, with minimal directional error. This process is repeated, extending the edge polylines progressively until no further growth is possible.
To ensure efficiency and accuracy, a non-maximum suppression step is employed to remove any redundant points that may exist between the current candidate and the best-matched point.
Please refer to the supplements for more algorithm details.

\begin{figure*}[!t]
\centering
\scalebox{0.9}{
\setlength\tabcolsep{2pt} %
\begin{tabular}{@{}cccccc@{}}
CAD model & LIMAP@S~\cite{liu20233d} & NEF@P~\cite{ye2023nef} & Ours@P &Ours@D & GT Edge \\
\includegraphics[width=2.5cm]{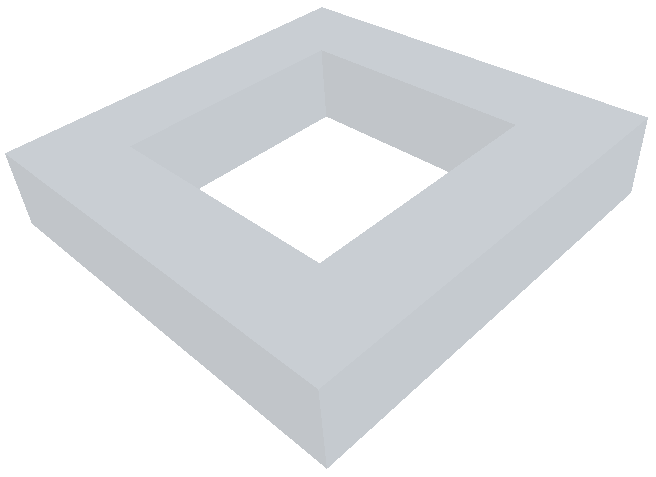} & \includegraphics[width=2.5cm]{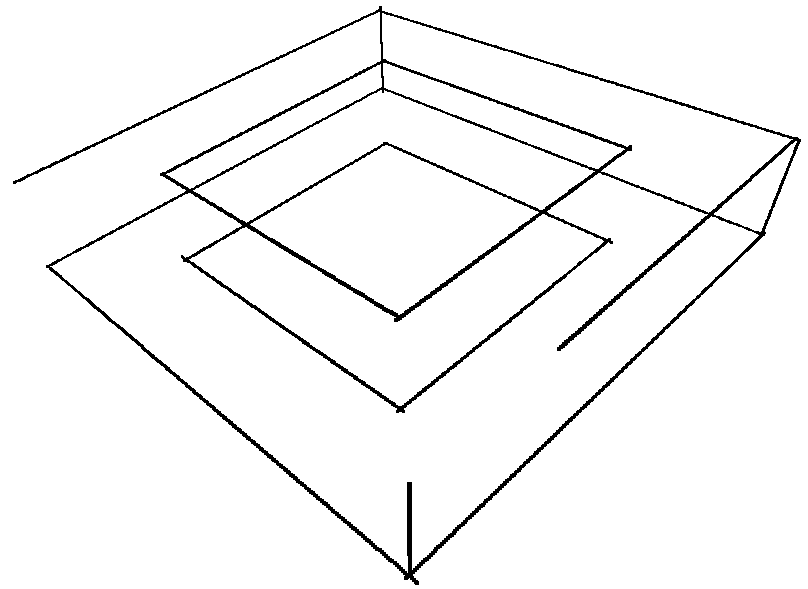} & \includegraphics[width=2.5cm]{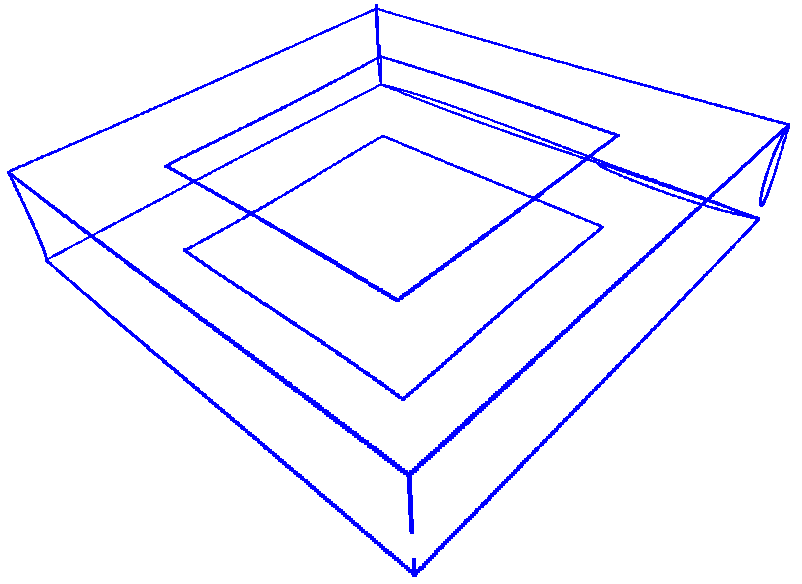} & \includegraphics[width=2.5cm]{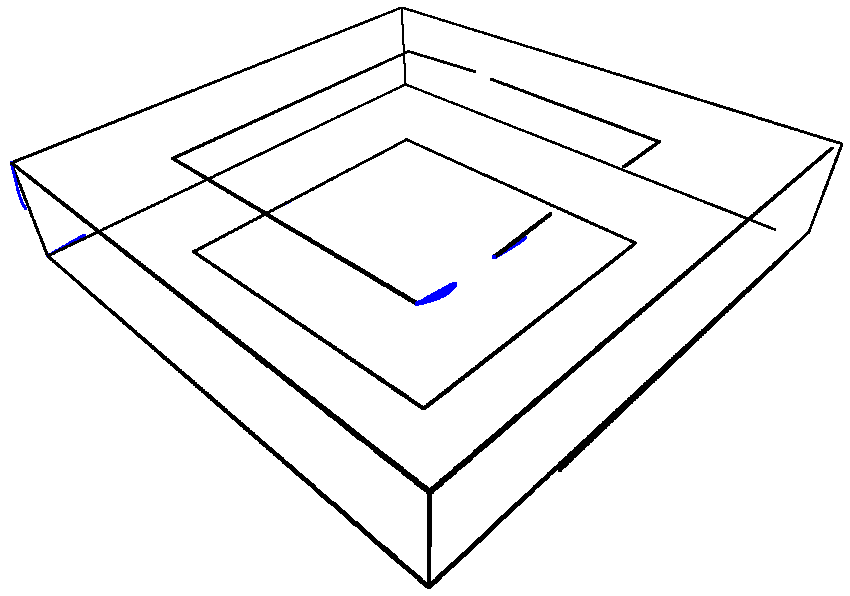} &
\includegraphics[width=2.5cm]{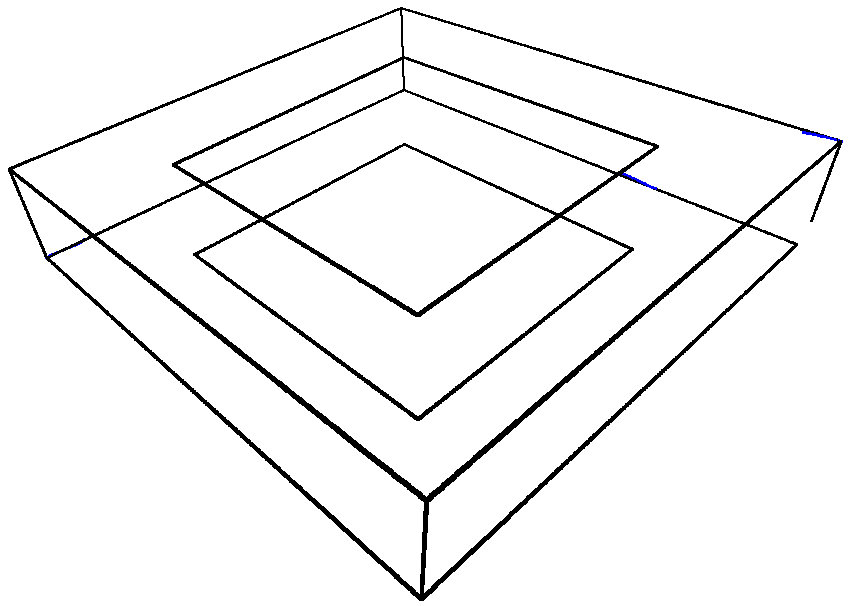} &\includegraphics[width=2.5cm]{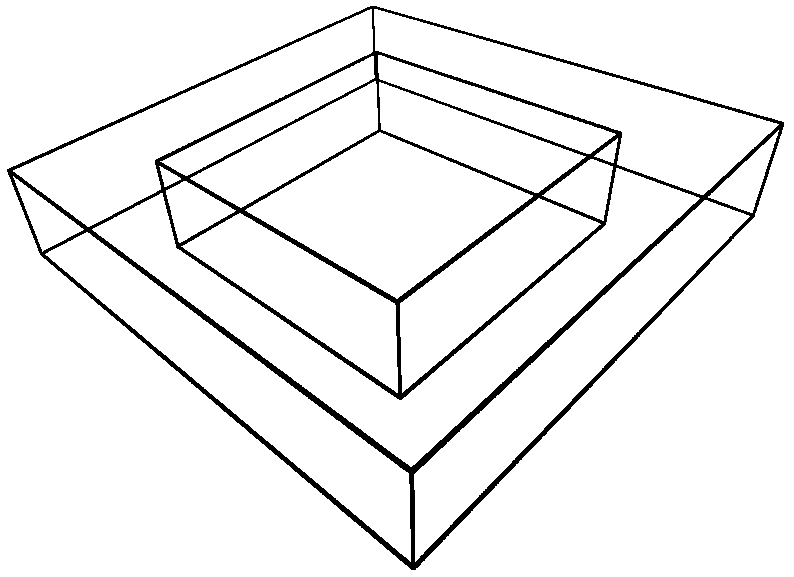} \\
\includegraphics[width=2.5cm]{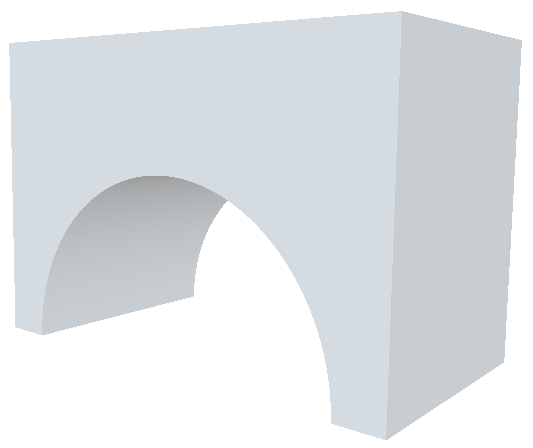} & \includegraphics[width=2.5cm]{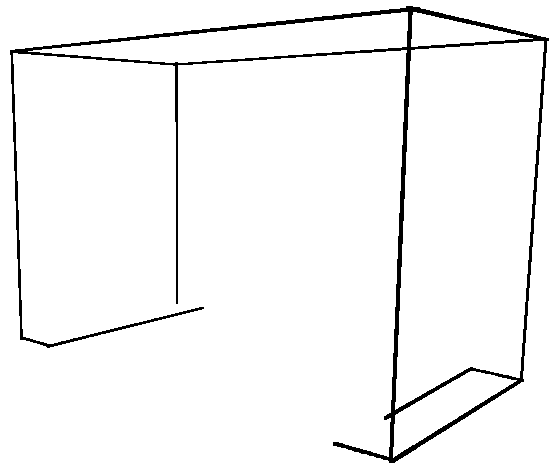} & \includegraphics[width=2.5cm]{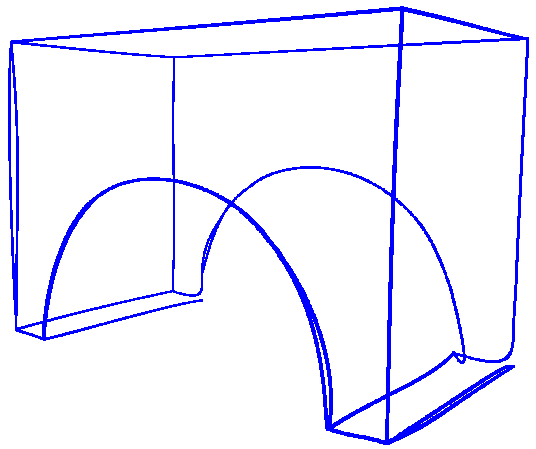} &
\includegraphics[width=2.5cm]{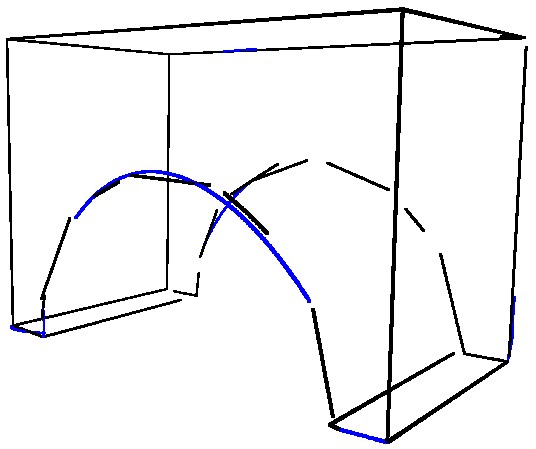} &
\includegraphics[width=2.5cm]{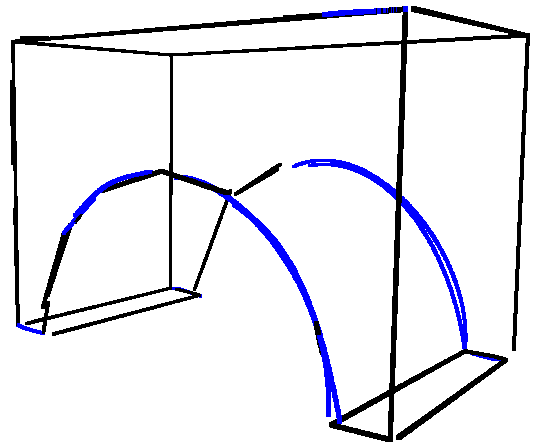} & \includegraphics[width=2.5cm]{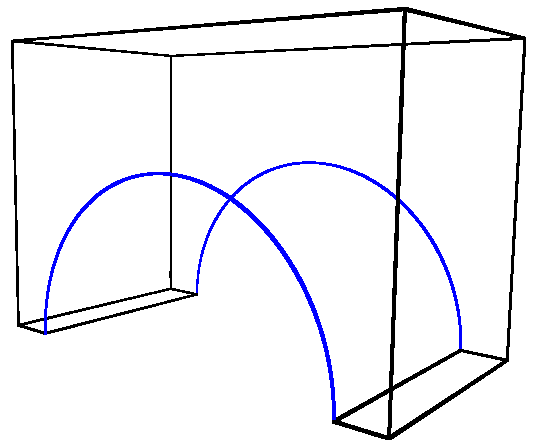} \\
\includegraphics[width=2.5cm]{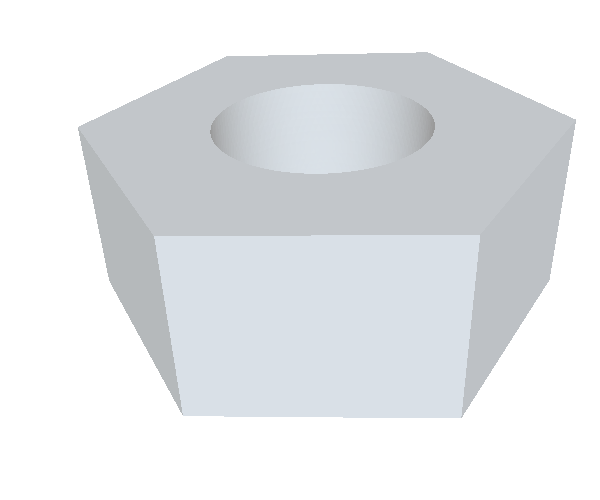} & \includegraphics[width=2.5cm]{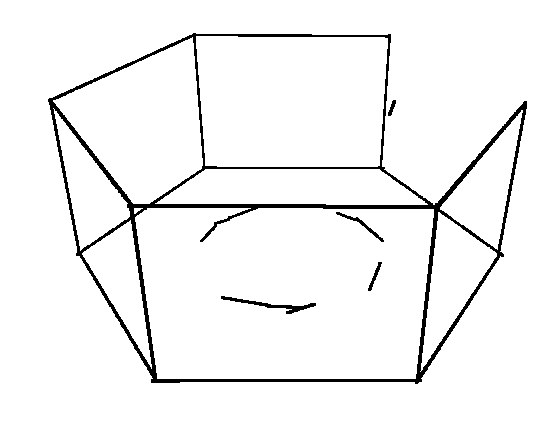} & \includegraphics[width=2.5cm]{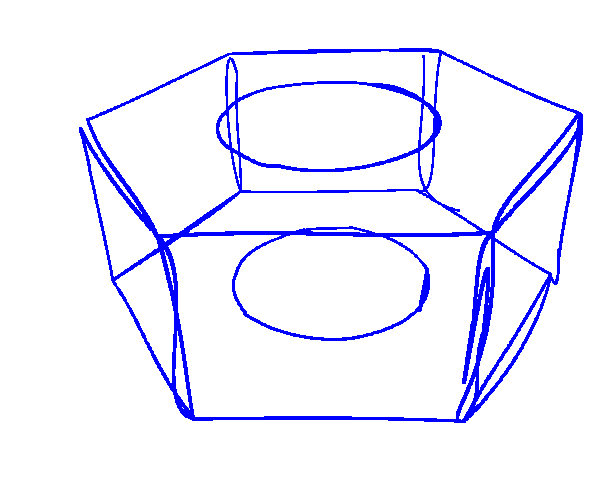} & \includegraphics[width=2.5cm]{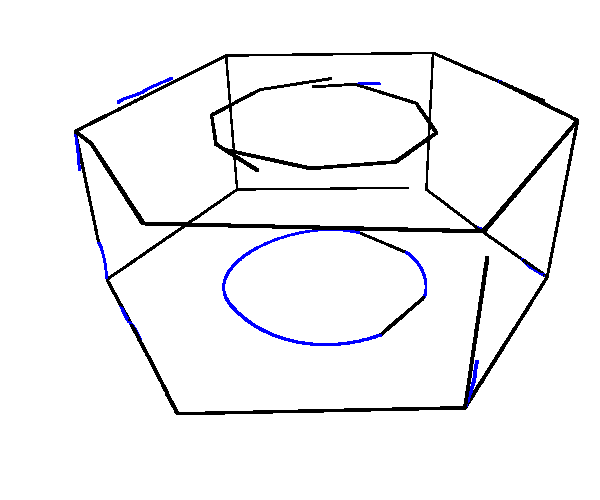} & 
\includegraphics[width=2.5cm]{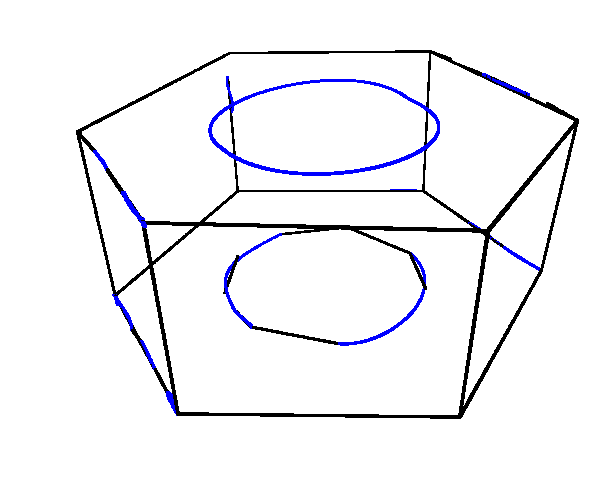} &
\includegraphics[width=2.5cm]{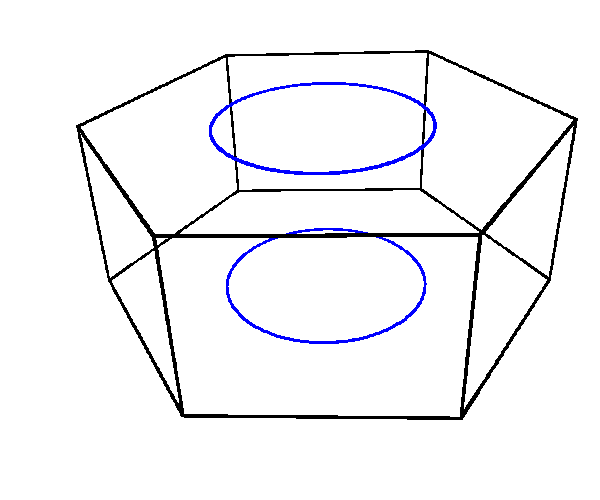}
\end{tabular}
}
\vspace{-1.0em}
\caption{\textbf{Qualitative comparisons on ABC-NEF~\cite{ye2023nef}.} Lines are shown in black and curves in blue. Thanks to our precise edge extraction capabilities for both lines and curves, we achieve complete and accurate modeling of these elements.}
\label{fig:ABC_comp}
\end{figure*}

\begin{table*}[!t]
\centering
\scalebox{0.8}{
\begin{tabular}{c|c|c|ccc|ccc|ccc|ccc}

  Method &
  Detector &
  Modal &
  Acc$\downarrow$ &
  Comp$\downarrow$ &
  Norm$\uparrow$ &
  R5$\uparrow$ &
  R10$\uparrow$ &
  R20$\uparrow$ &
  P5$\uparrow$ &
  P10$\uparrow$ &
  P20$\uparrow$ &
  F5$\uparrow$ &
  F10$\uparrow$ &
  F20$\uparrow$ \\ \hline
  \multirow{2}{*}{LIMAP~\cite{liu20233d}} &
  LSD &
  Line &
  9.9 &
  18.7 &
  94.4 &
  36.2 &
  82.3 &
  87.9 &
  43.0 &
  87.6 &
  93.9 &
  39.0 &
  84.3 &
  90.4 \\

   &
  SOLD2 &
  Line &
  \textbf{5.9} &
  29.6 &
  90.1 &
  \textbf{64.2} &
  76.6 &
  79.6 &
  \textbf{88.1} &
  \textbf{96.4} &
  \textbf{97.9} &
  \textbf{72.9} &
  84.0 &
  86.7 \\ \hline
  \multirow{3}{*}{NEF~\cite{ye2023nef}} &
  PiDiNet$\dagger$ &
  Curve &
  11.9 &
  16.9 &
  90.9 &
  11.4 &
  62.0 &
  91.3 &
  15.7 &
  68.5 &
  96.3 &
  13.0 &
  64.6 &
  93.3 \\
   &
  PiDiNet &
  Curve &
  15.1 &
  16.5 &
  89.7 &
  11.7 &
  53.3 &
  89.8 &
  13.6 &
  52.2 &
  89.1 &
  12.3 &
  51.8 &
  88.7 \\
   &
  DexiNed &
  Curve &
  21.9 &
  15.7 &
  85.9 &
  11.3 &
  48.3 &
  87.7 &
  11.5 &
  39.8 &
  71.7 &
  10.8 &
  42.1 &
  76.8 \\ \cline{1-15} 
  \multirow{2}{*}{\textbf{Ours}} &
  PiDiNet &
  Edge &
  9.2 &
  15.6 &
  93.7 &
  30.2 &
  75.7 &
  89.8 &
  35.6 &
  79.1 &
  95.4 &
  32.4 &
  77.0 &
  92.2 \\
  &
  DexiNed &
  Edge &
  8.8 &
  \textbf{8.9} &
  \textbf{95.4} &
  56.4 &
  \textbf{88.9} &
  \textbf{94.8} &
  62.9 &
  89.9 &
  95.7 &
  59.1 &
  \textbf{88.9} &
  \textbf{94.9}
\end{tabular}
}
\vspace{-0.5em}
\caption{\textbf{Edge reconstruction results on ABC-NEF~\cite{ye2023nef}.} Results from NEF's released pretrained models are indicated by $\dagger$. Our method surpasses all others in terms of completeness and achieves accuracy comparable to LIMAP~\cite{liu20233d}.
}

\label{label:ABC}
\vspace{-1.em}
\end{table*}

\boldparagraph{Edge Fitting} 
To further parameterize edges, we categorize the polylines into line segments and Bézier curves (Fig. \ref{fig:postprocessing}~(f)). 
Initially, we utilize RANSAC~\cite{fischler1981random} to fit lines from the polylines, and select the line segment that encompasses the highest number of inlier points. Following \cite{liu20233d}, we apply Principal Component Analysis (PCA) to the inlier points, re-estimate the line segment utilizing the principal eigenvector and the mean 3D point, and project all inlier points onto the principal eigenvector to derive the 3D endpoints.  
This fitting process is repeated for each polyline until the number of inlier points falls below a minimum threshold.
For the remaining sub-polylines, we fit each of them with a Bézier curve that is defined by four control points.

To minimize edge redundancy, we further merge line segments and Bézier curves based on two criteria: the shortest distance between candidate edges and the similarity of curvature at their closest points.
For line segments, the shortest distance is the minimal point-to-line segment distance, and curvature similarity is their direction's cosine similarity. For Bézier curves, they are the minimal point-to-point distance and the cosine similarity of the tangent vectors at the nearest points, respectively.
Candidate edges are merged only if they meet both criteria. This dual-criterion approach ensures that merging happens only when two edges are both similar and close to each other.

To connect edges, all endpoints of line segments and Bézier curves located within a specified distance threshold are merged into shared endpoints. 
Furthermore, we implement an optimization step~\cite{bignoli2018edgegraph3d, xue2023volumetric} to refine the 3D parametric edges by leveraging 2D edge maps, thereby enhancing edge precision. Specifically, we project 3D parametric edges into edge map frames using camera projection matrices and filter out 3D edges that are not visible in over 90\% of views.

\section{Experiments}
\subsection{Experiment Setting}
\boldparagraph{Datasets} We consider four diverse datasets: CAD models (ABC-NEF~\cite{ye2023nef}), real-world objects (DTU~\cite{aanaes2016large}), high-quality indoor scenes (Replica~\cite{straub2019replica}), and real-world outdoor scenes (Tanks \& Temples~\cite{knapitsch2017tanks}). 
ABC-NEF dataset comprises 115 CAD models, each accompanied by 50 observed images and ground truth parametric edges. We select 82 CAD models, excluding those containing inconsistent edge observations (e.g., cylinders or balls).
DTU dataset provides dense ground-truth point clouds and we select 6 objects that meet the multi-view constraints among scans processed by \cite{yu2022monosdf}. Following \cite{bignoli2018edgegraph3d}, we derive edge points by projecting ground-truth dense points onto images and then comparing them with the observations on 2D edge maps to filter out non-edge points. 
Replica and Tanks \& Temples datasets contain larger scenes. Due to the lack of ground-truth edges, we conduct qualitative comparisons among baselines.

\boldparagraph{Baselines} We compare with three state-of-the-art baselines for 3D line/curve mapping, including two learning-based methods, NEF~\cite{ye2023nef} and NEAT~\cite{xue2023volumetric}, and one geometry-based method, LIMAP~\cite{liu20233d}.

\boldparagraph{Metrics} 
Our evaluation involves first sampling points in proportion to the edge's length and subsequently downsampling these points using a voxel grid with a resolution of $256^3$.
Following the metrics used in \cite{ye2023nef, liu20233d}, we consider Accuracy (Acc),  Completeness (Comp) in millimeters, and Recall ($R_\tau$), Precision ($P_\tau$), F-score ($F_\tau$) in percentage with a threshold $\tau$ in millimeters. Moreover, we report Edge Direction Consistency (Norm) in percentage to analyze the precision of edge direction extraction.

\boldparagraph{Implementation Details} 
For $f_u$, we utilize 8-layer Multilayer Perceptrons (MLPs). Each layer in the MLP contains 512 neurons for larger scenes, such as Tanks \& Temples, and 256 neurons for other datasets.
We sample 1024 rays per batch, among these rays, 512 rays are sampled from edge areas. We train our model for $50k$ iterations on ABC-NEF dataset, and $200k$ iterations on other datasets. 
We train our network with the Adam optimizer with a learning rate of $5 \times 10^{-4}$, while the UDF model $f_u$ is trained with a learning rate of $1 \times 10^{-4}$ and initialized with sphere initialization~\cite{Yariv2020NEURIPS}.
For edge detection for NEF and ours, we consider PiDiNet~\cite{su2021pixel} and DexiNed~\cite{poma2020dense}.
PiDiNet~\cite{su2021pixel} is employed for indoor scenes, such as DTU and Replica, due to its superior performance in these settings. Conversely, DexiNed~\cite{poma2020dense} is applied to outdoor scenes, as it is primarily trained on outdoor scenes.
On the synthetic ABC-NEF dataset, we show results with both detectors. 
For LIMAP, we follow their paper and we use SOLD2~\cite{pautrat2021sold2} for indoor scenes and LSD~\cite{von2008lsd} for outdoor scenes. 
NEAT is trained with 2D wireframes from HAWPV3~\cite{xue2023holistically}.

\begin{figure*}[ht]
\centering
\vspace{-0.5em}
\scalebox{0.77}{
\setlength\tabcolsep{2pt} %
\begin{tabular}{cccc}
\addlinespace[5pt]
\\
2D Image & LIMAP~\cite{liu20233d} & NEAT~\cite{xue2023volumetric} & Ours \\
\includegraphics[width=4.4cm]{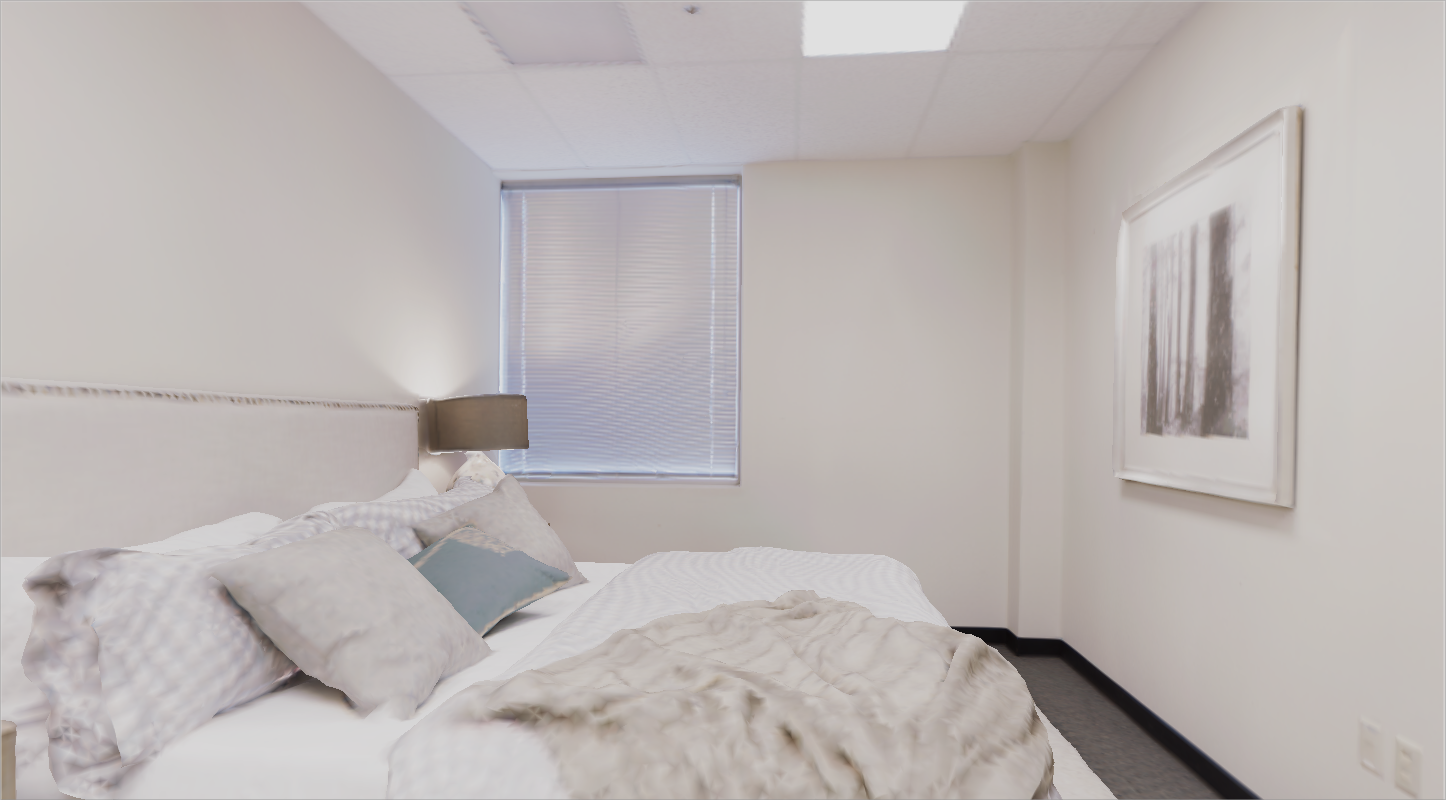}
& \includegraphics[width=4.4cm]{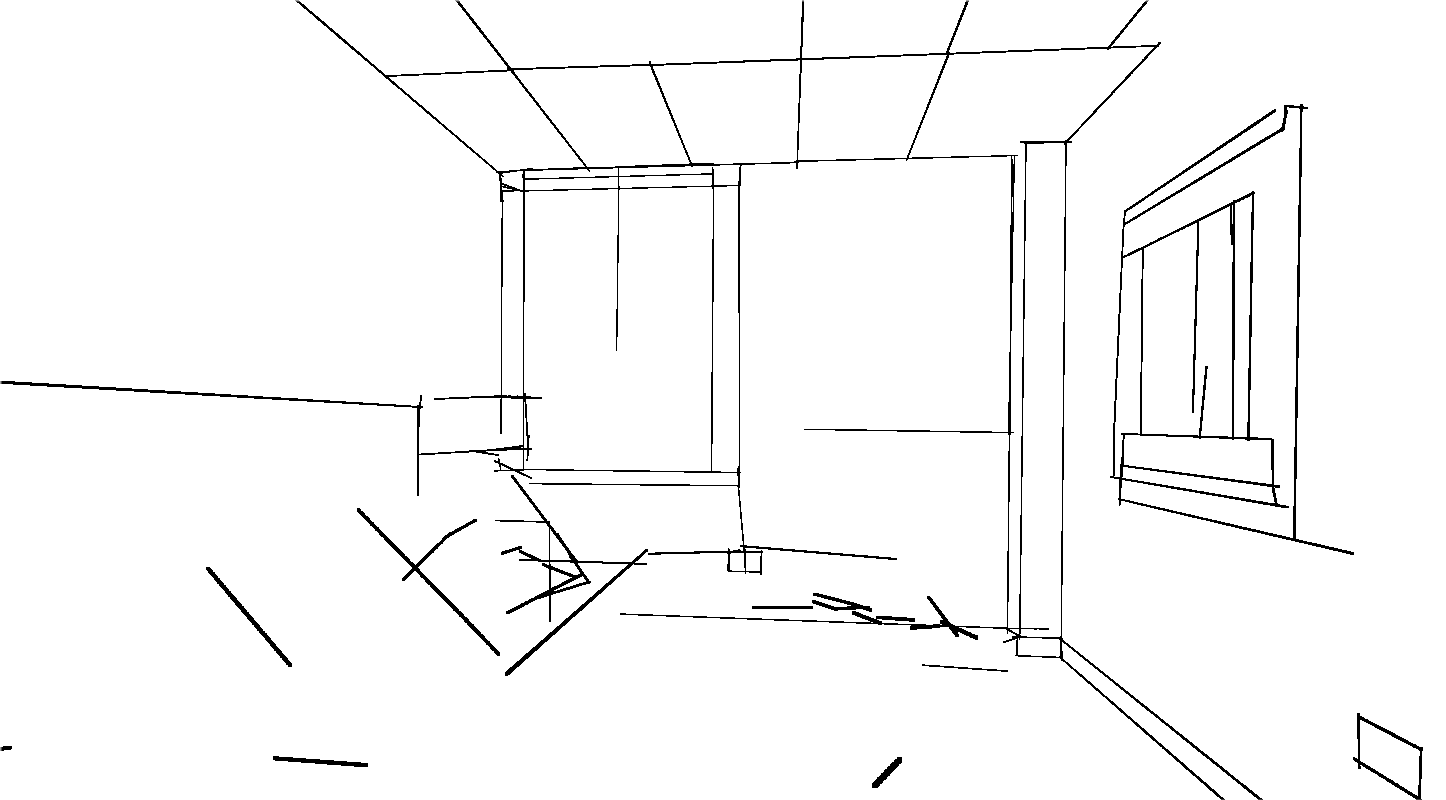} & \includegraphics[width=4.4cm]{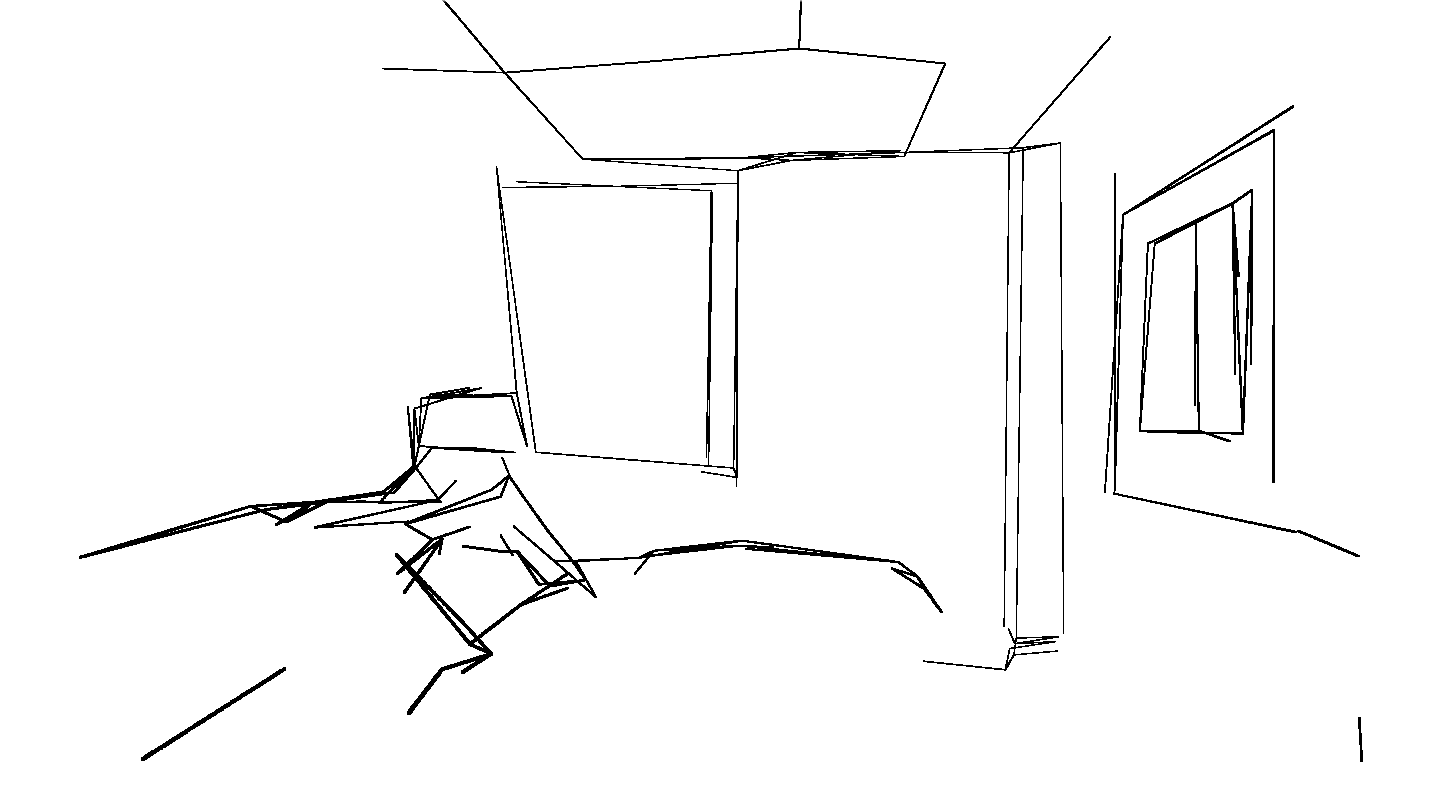}
& \includegraphics[width=4.4cm]{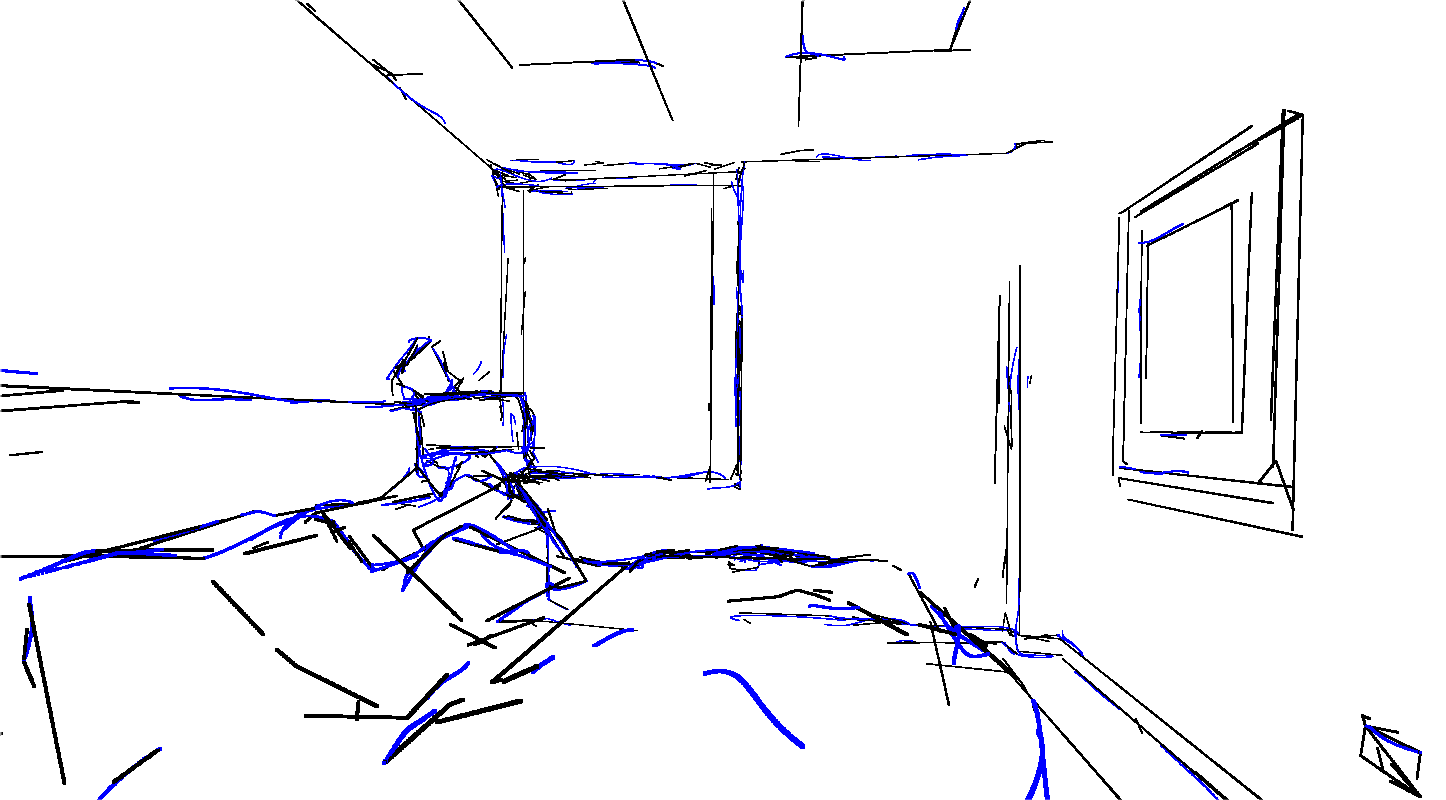}\\
\includegraphics[width=4.4cm]{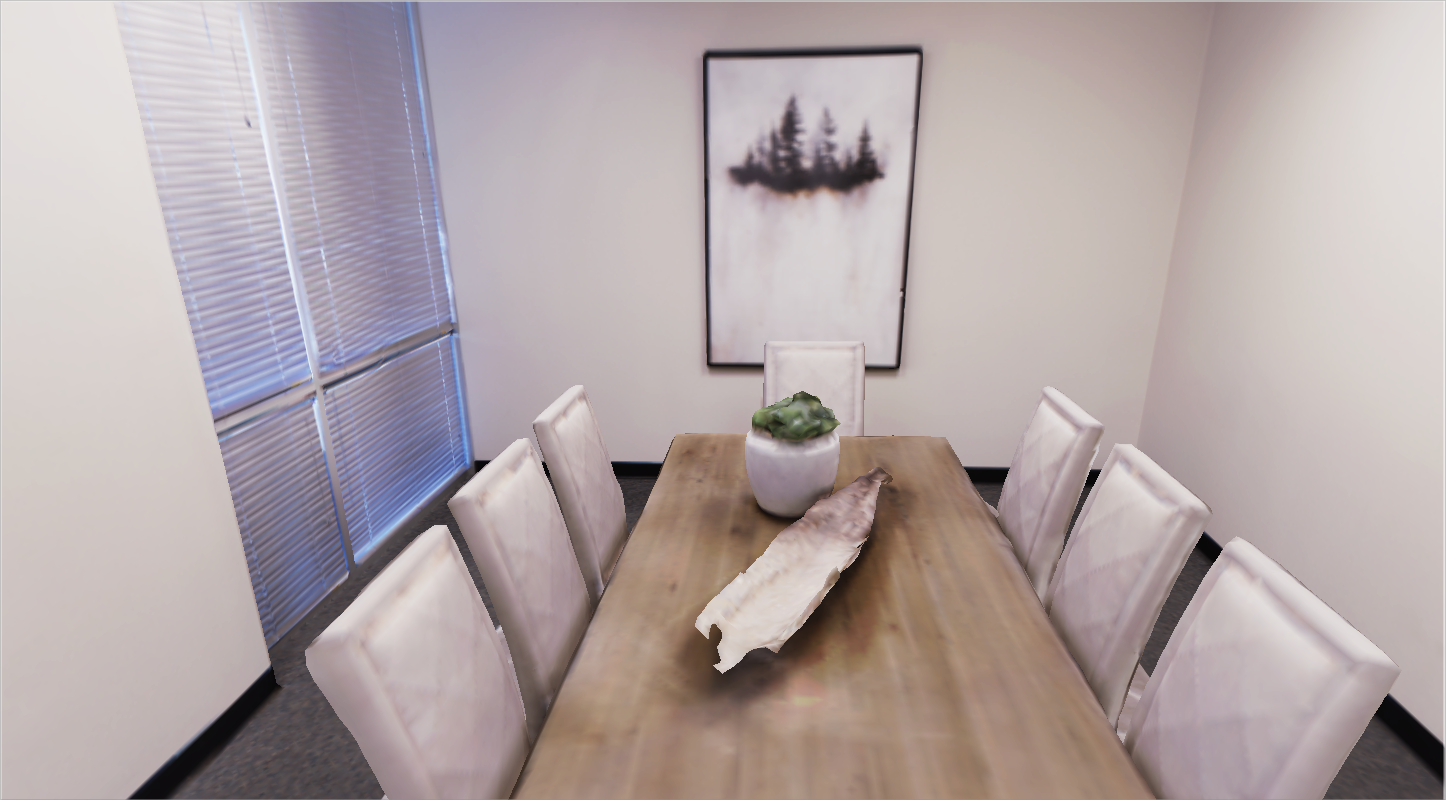}
& \includegraphics[width=4.4cm]{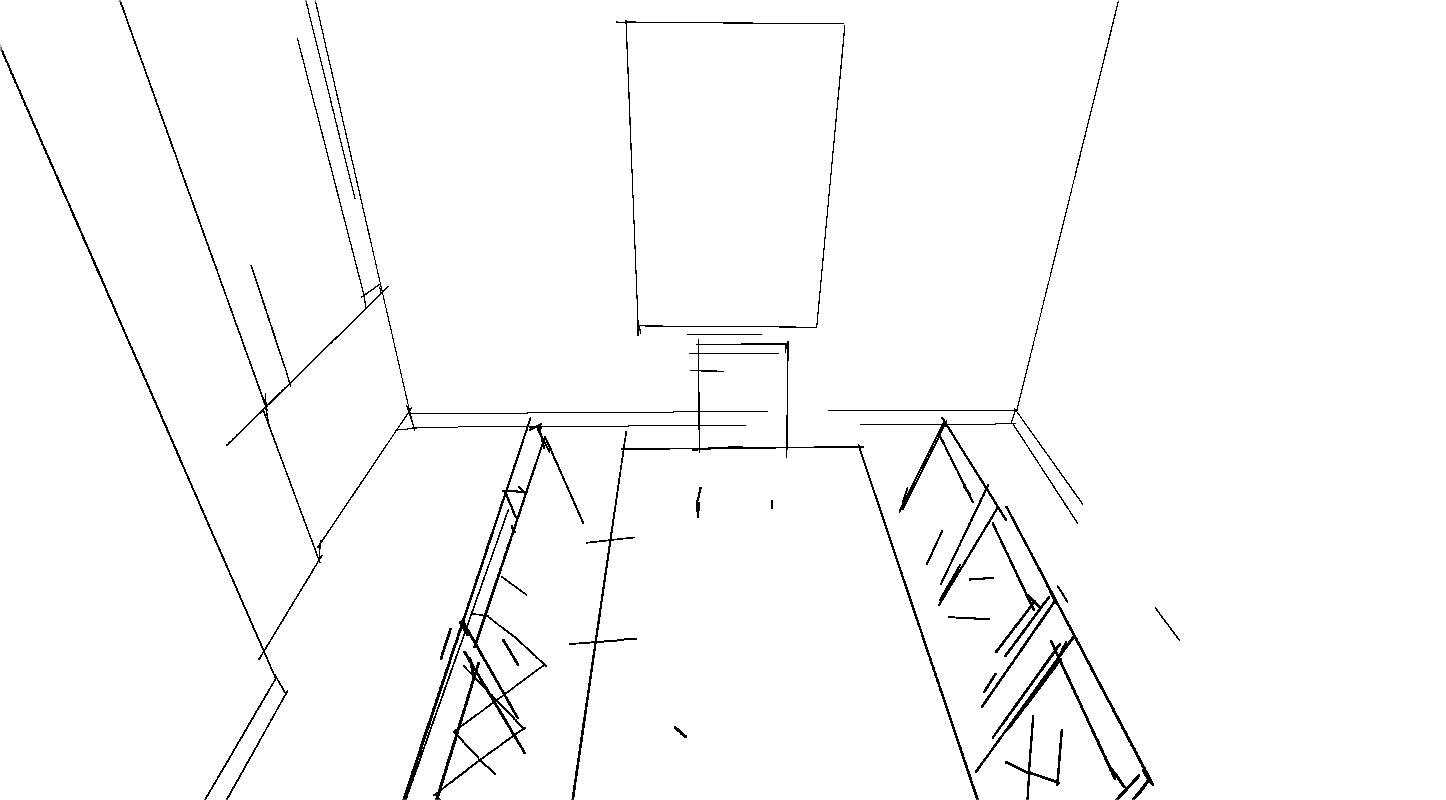} & \includegraphics[width=4.4cm]{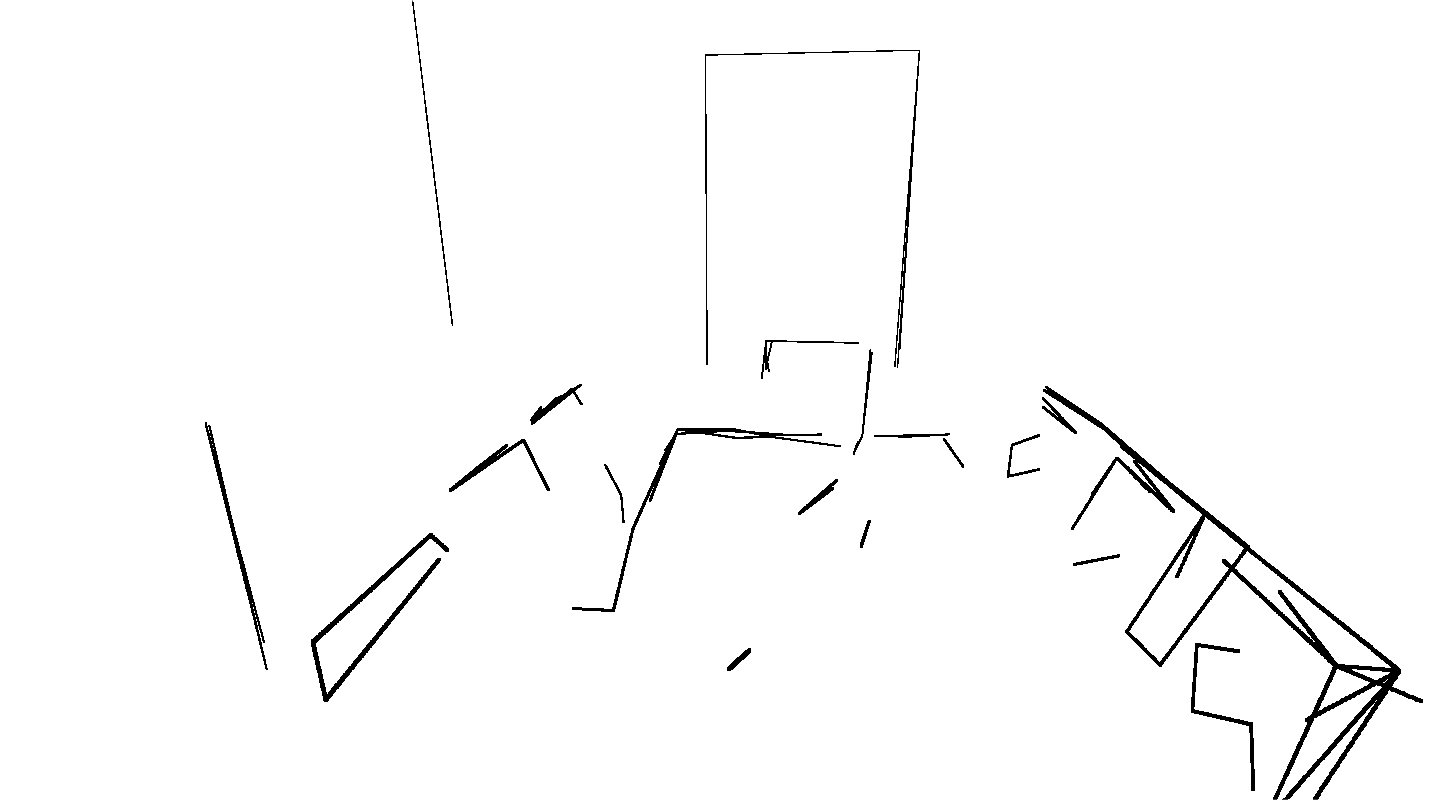}
& \includegraphics[width=4.4cm]{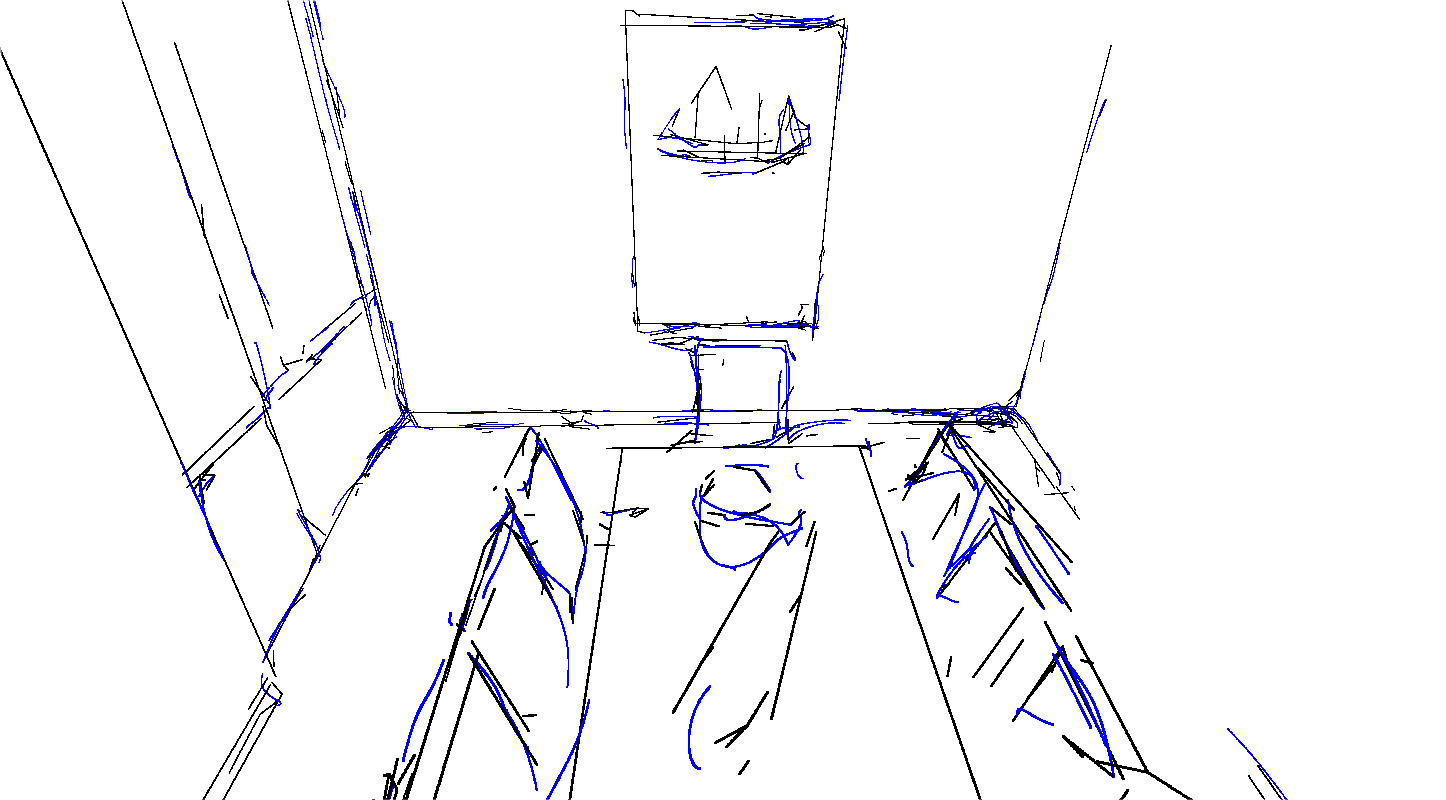} \\
\includegraphics[width=4.4cm]{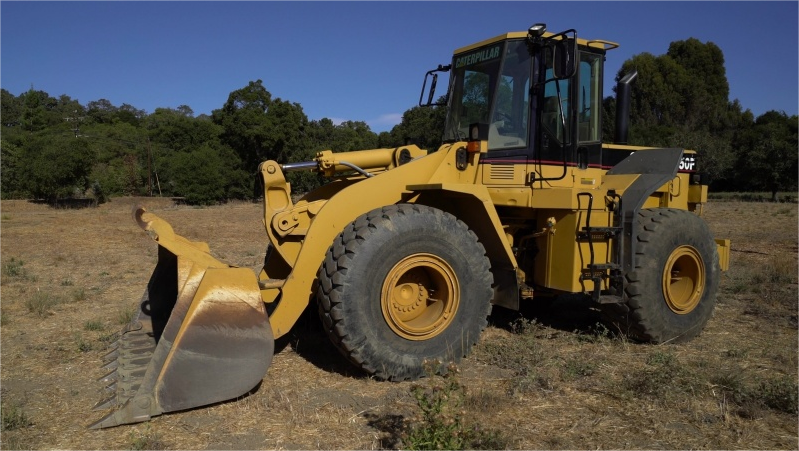}
& \includegraphics[width=4.4cm]{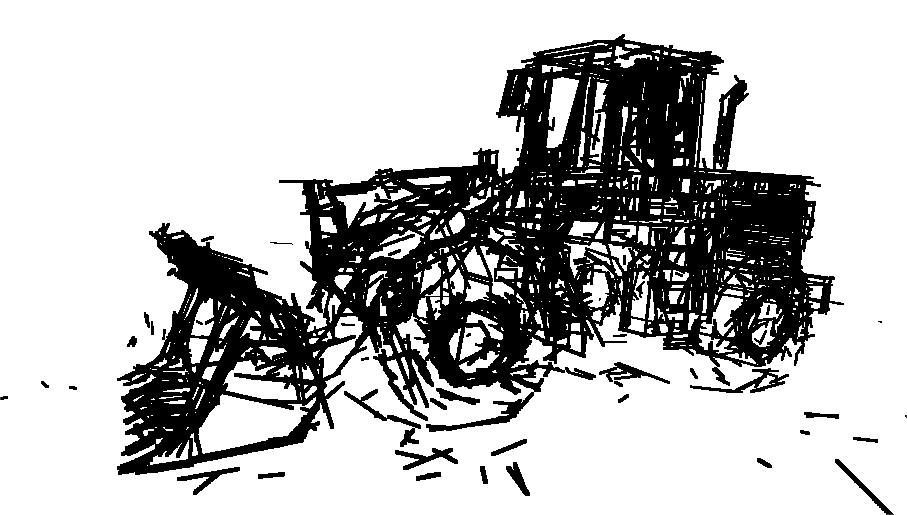} & \includegraphics[width=4.4cm]{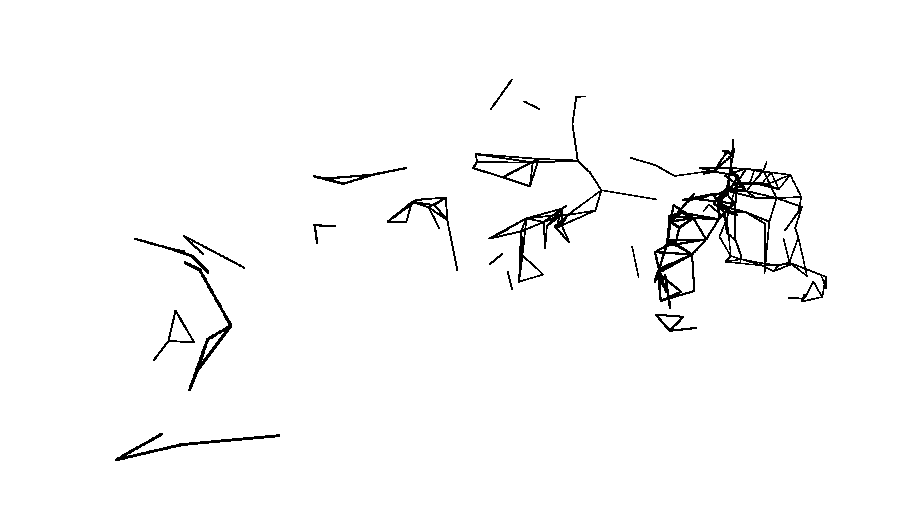}
& \includegraphics[width=4.4cm]{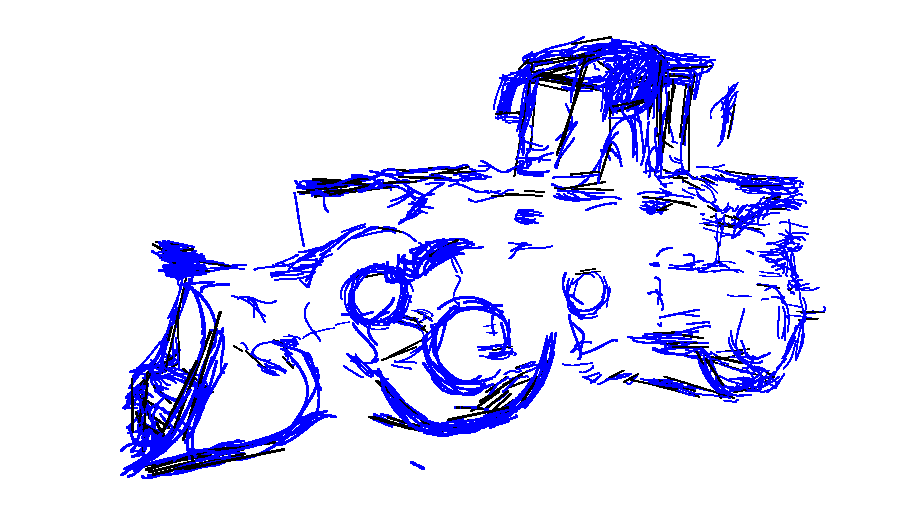} \\
\end{tabular}
}
\vspace{-0.5em}
\caption{\textbf{Qualitative comparisons on the Replica~\cite{straub2019replica} and Tanks \& Temples~\cite{knapitsch2017tanks} datasets.} The first two scenes are from the Replica dataset, while the last scene is from the Tanks \& Temples dataset.
}
\label{fig:replica_blendedmvs}
\vspace{-1.0em}
\end{figure*}

\subsection{Evaluation of 3D Edge Reconstruction}
\boldparagraph{Evaluation on ABC-NEF Dataset} 
We show the quantitative and qualitative comparisons on~\tabref{label:ABC} and~\figref{fig:ABC_comp}. Note that NEAT fails on the ABC-NEF dataset because of its heavy dependence on texture input.
NEF demonstrates decent performance at $\tau=20$. However, their performance drops significantly when $\tau$ is set to 10 and 5. 
This is attributed to its bias in edge rendering and its fitting-based post-processing. 
LIMAP shows remarkable precision across various $\tau$ thresholds. Such consistency stems from its non-linear refinement over multi-view 2D supports. Nonetheless, LIMAP's inability to reconstruct curves leads to lower scores in completeness and recall.
Our method, when combined with either of the 2D edge detectors, consistently outperforms all baselines.
Notably, as shown in~\tabref{label:ABC}, combined with the DexiNed detector, our method achieves superior results in completeness, edge direction consistency, recall, and F-Score. We also show competitive accuracy and precision when compared to LIMAP.

\begin{table}[t]
\centering
\scalebox{0.65}{

\begin{tabular}{c|c|ccc|ccc}

\multirow{2}{*}{Method} &
  \multirow{2}{*}{Detector} &
  \multicolumn{3}{c|}{Curve} &
  \multicolumn{3}{c}{Line} \\
 & &
   Acc$\downarrow$ &
  Comp$\downarrow$ &
  Norm$\uparrow$ &
  Acc$\downarrow$ &
  Comp$\downarrow$ &
  Norm$\uparrow$\\ \hline
\multirow{2}{*}{LIMAP~\cite{liu20233d}} & LSD     & 272.6 & 50.1 & 84.8  & 34.6 & 11.3 & 95.9 \\
                       & SOLD2 & 295.7 & 82.2 & 76.8 & \textbf{20.0} & 18.1 & 92.1 \\ \hline
\multirow{3}{*}{NEF~\cite{ye2023nef}}   & PiDiNet$\dagger$ & 265.0 & 27.1 & 77.9 & 40.4 & 13.7 & 92.6\\

& PiDiNet & 263.1 & 23.9 & 77.6 & 43.9 & 14.0 & 91.4 \\
                       & DexiNed & 250.5 & 20.3 & 72.6 & 56.2 & 13.8 & 87.3 \\ \hline
\multirow{2}{*}{\textbf{Ours}}  & PiDiNet & 253.7 & 25.7 & 88.1 & 43.1 & 12.8 & 93.7\\
 &
  DexiNed & \textbf{241.0} & \textbf{10.9}  & \textbf{88.7} & 46.7 & \textbf{7.7} & \textbf{95.4} \\

\end{tabular}
}
\vspace{-0.5em}

\caption{\textbf{Accuracy, completeness and normal consistency results with curves and lines on ABC-NEF~\cite{ye2023nef}.} Our method with DexiNed edge detector yields overall the strongest performance on curves among all baselines.
}
\label{label:ABC_edge_type}
\vspace{-1.8em}
\end{table}

To further analyze the performance of different edge types, we classify the ground truth edges into curves (including BSplines, ellipses, and circles) and line segments, based on the GT annotations.
We provide accuracy, completeness, and edge direction consistency in~\tabref{label:ABC_edge_type} to analyze the separate reconstruction abilities for curves and lines.
Note that these results are computed based on all predictions specific to either curves or lines, as other methods do not differentiate between these two types of edges.
We can see that our method with DexiNed exhibits superior results in reconstructing curves. 
As for line segments, our performance is marginally lower than the best-performing method LIMAP which is specially optimized for lines.

\begin{figure}[!t]
\centering
\setlength\tabcolsep{2pt} %
\scalebox{0.85}{
\begin{tabular}{@{}cccc@{}}
\includegraphics[width=2.8cm]{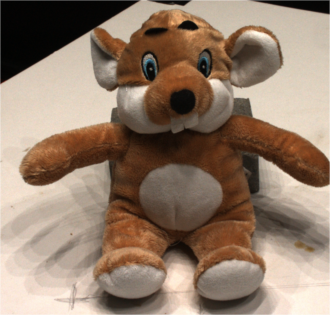} & \includegraphics[width=2.8cm]{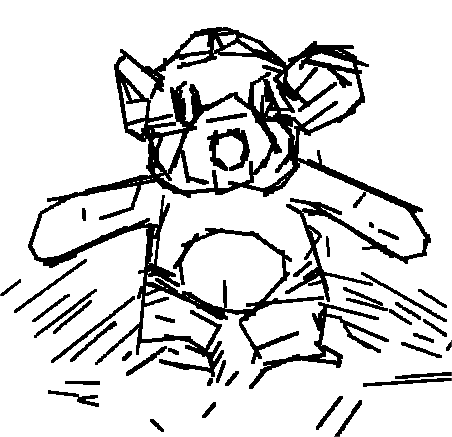} & \includegraphics[width=2.8cm]{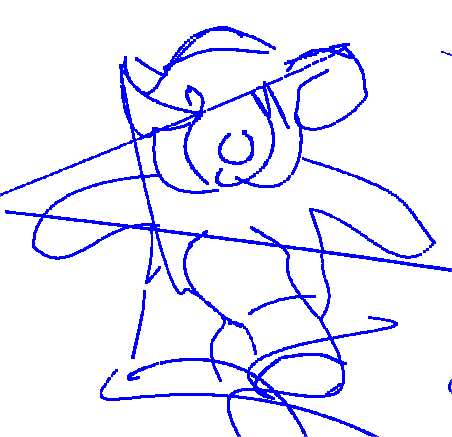} \\
2D Image & LIMAP~\cite{liu20233d} & NEF~\cite{ye2023nef}\\
\includegraphics[width=2.8cm]{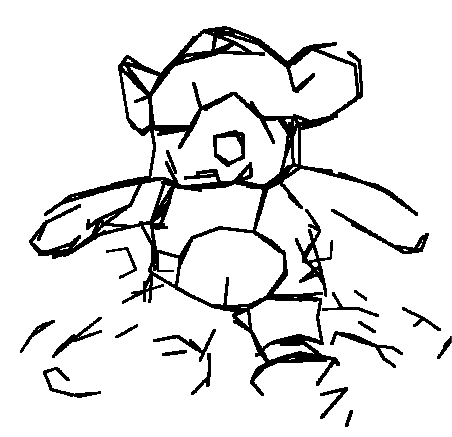} &
\includegraphics[width=2.8cm]{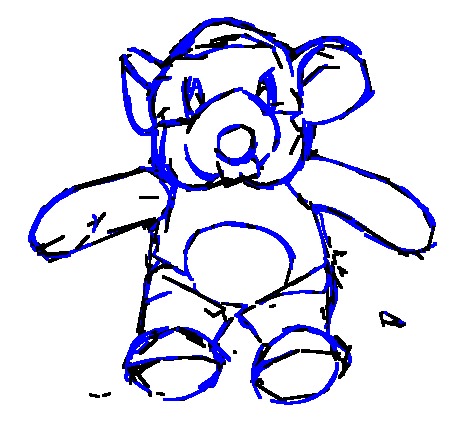} &
\includegraphics[width=2.8cm]{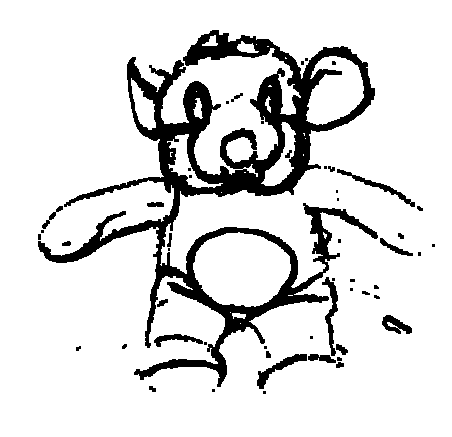} \\
NEAT~\cite{xue2023volumetric} & Ours & GT Edge \\
\end{tabular}
}
\vspace{-0.5em}
\caption{\textbf{Qualitative comparisons on DTU~\cite{aanaes2016large}.} Our results demonstrate complete edge structure, whereas other methods result in redundant line segments or imprecise curves.}
\label{fig:dtu}
\vspace{-0.5em}
\end{figure}

\begin{table}[!t]
\centering
\scalebox{0.75}{

\begin{tabular}{c|cc|cc|cc|cc}

\multirow{2}{*}{Scan} & \multicolumn{2}{c|}{LIMAP~\cite{liu20233d}}  & \multicolumn{2}{c|}{NEF~\cite{ye2023nef}}    & \multicolumn{2}{c|}{NEAT~\cite{xue2023volumetric}}   & \multicolumn{2}{c}{\textbf{Ours}}    \\
                      & R5$\uparrow$ &
  P5$\uparrow$ &
  R5$\uparrow$ &
  P5$\uparrow$ &
  R5$\uparrow$ &
  P5$\uparrow$ &
  R5$\uparrow$ &
  P5$\uparrow$ \\ \hline
37   & \textbf{75.8} & 74.3 & 39.5 & 51.0 & 63.9 & \textbf{85.1} & 62.7          & 83.9          \\
83   & \textbf{75.7} & 50.7 & 32.0 & 21.8 & 72.3 & 52.4          & 72.3          & \textbf{61.5} \\
105  & \textbf{79.1} & 64.9 & 30.3 & 32.0 & 68.9 & 73.3          & 78.5          & \textbf{78.0} \\
110  & 79.7          & 65.3 & 31.2 & 40.2 & 64.3 & \textbf{79.6} & \textbf{90.9} & 68.3          \\
118  & 59.4          & 62.0 & 15.3 & 25.2 & 59.0 & 71.1          & \textbf{75.3} & \textbf{78.1} \\
122  & 79.9          & 79.2 & 15.1 & 29.1 & 70.0 & 82.0          & \textbf{85.3} & \textbf{82.9} \\ \hline
Mean & 74.9          & 66.1 & 27.2 & 33.2 & 66.4 & 73.9          & \textbf{77.5} & \textbf{75.4}
\end{tabular}
}
\vspace{-0.5em}
\caption{\textbf{Edge reconstruction results on DTU~\cite{aanaes2016large}}.} 
\label{tab:dtu}
\vspace{-1.5em}
\end{table}

\boldparagraph{Evaluation on DTU Dataset}
Our assessment of the DTU dataset, as outlined in~\tabref{tab:dtu} and~\figref{fig:dtu}, shows our proficiency in real-world scenarios. Notably, our approach achieves the highest recall and precision among all baselines. 
The DTU dataset presents a challenging scenario for edge extraction due to its varying lighting conditions. However, our edge refinement step proves effective in preserving primary edges, a point we elaborate on in \secref{sec:edge_refine}.
\figref{fig:dtu} shows LIMAP tends to produce redundant line segments, leading to high recall but reduced precision. 
NEF's post-processing is sensitive to different scenes, resulting in noisy edge fitting. 
NEAT, despite producing clean outputs, its inability to handle curves constrains its overall performance.

\boldparagraph{Qualitative Evaluation on Indoor \& Outdoor Scenes} 
To really showcase the power of our method in capturing scene-level geometry, we further run our method on indoor and outdoor scenes. 
Note that since NEF is not able to produce meaningful reconstructions on larger scenes, we only compare with LIMAP and NEAT.
As shown in ~\figref{fig:teaser} and ~\figref{fig:replica_blendedmvs}, NEAT, due to its reliance on high-quality surface reconstruction, faces limitations in scene reconstruction, while LIMAP and our method both successfully capture good scene geometry.

\subsection{Ablations and Analysis}
\begin{figure}[!t]
    \centering
    \subfloat[w/o point shifting]{%
        \includegraphics[width=0.3\linewidth]{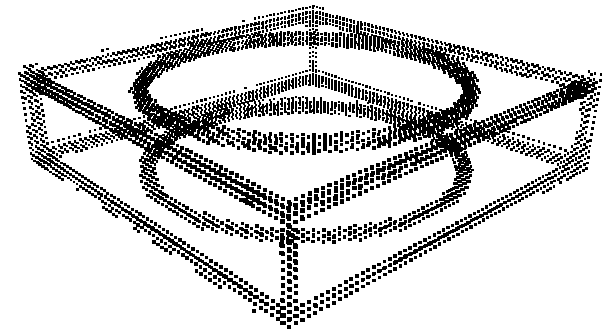}
        \label{fig:wo_ps}
    }
    \subfloat[w/ point shifting]{%
        \includegraphics[width=0.3\linewidth]{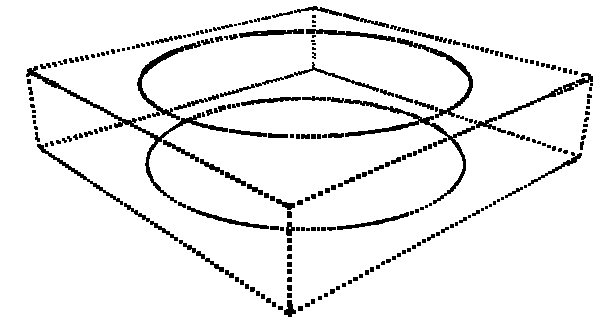}
        \label{fig:w_ps}
    }
    \subfloat[edge direction]{%
        \includegraphics[width=0.3\linewidth]{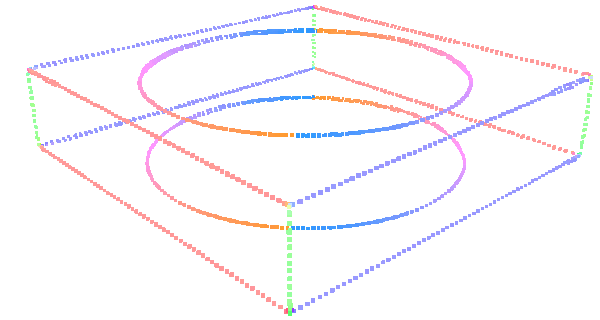}
        \label{fig:w_ps_direction}
    }
    \vspace{-0.5em}
    \caption{\textbf{Visualization of point shifting and edge direction.} Edge points are shown in point clouds and edge directions in color. The point shifting step significantly refines the locations of edge points. The edge extraction step yields accurate results, as seen in parallel lines sharing the same direction and curves exhibiting continuously changing directions.}
    \label{fig:point_shifting}
    \vspace{-1.5em}
\end{figure}

\boldparagraph{Parametric Edge Extraction}
To better understand our parametric edge extraction process described in~\secref{sec:edge_extraction}, we visualize our point shifting and edge direction in~\figref{fig:point_shifting}. 
We can clearly see that the extracted point clouds without point shifting are appeared in redundant and inaccurate edge points (\figref{fig:point_shifting}~(a)). In contrast, the point shifting step yields point clouds with sharply defined, precise edges (\figref{fig:point_shifting}~(b)). In addition, as shown in~\figref{fig:point_shifting}~(c), the extracted edge directions along parallel lines are consistent, while those on curves vary continuously. 
This aligns with our expectations.

\begin{figure}[!t]
\vspace{-0.5em}
    \centering
    \subfloat[Ours]{%
        \includegraphics[width=0.3\linewidth]{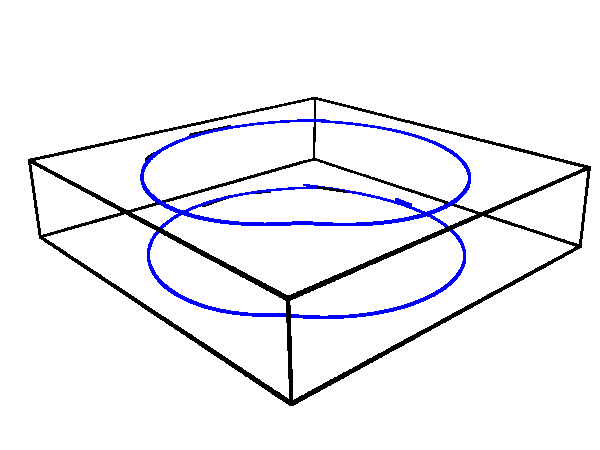}
        \label{fig:ablation_a}
    }
    \subfloat[w/o point shifting]{%
        \includegraphics[width=0.3\linewidth]{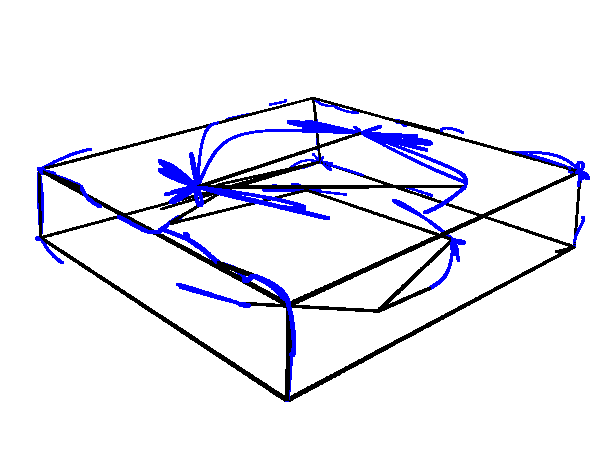}
        \label{fig:ablation_b}
    }
    \subfloat[w/o B\'ezier curve]{%
        \includegraphics[width=0.3\linewidth]{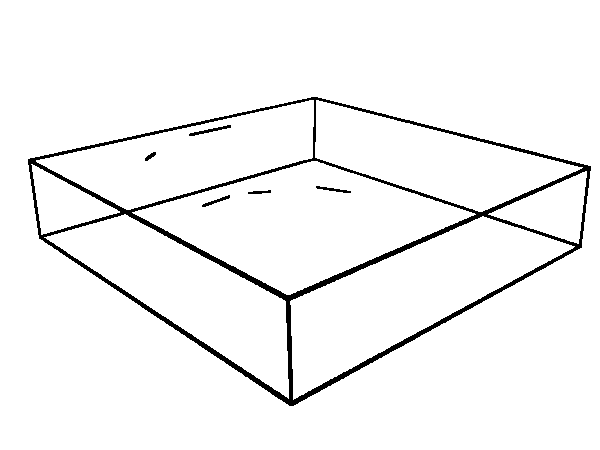}
        \label{fig:ablation_c}
    }\\
    \subfloat[w/o edge merg.]{%
        \includegraphics[width=0.3\linewidth]{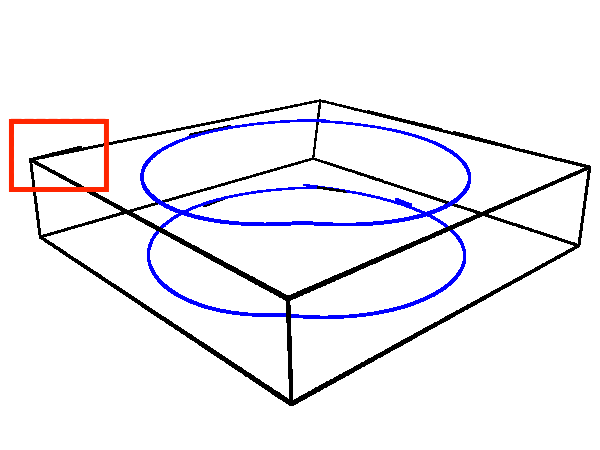}
        \label{fig:ablation_d}
    }
    \subfloat[w/o endpoints merg.]{%
        \includegraphics[width=0.3\linewidth]{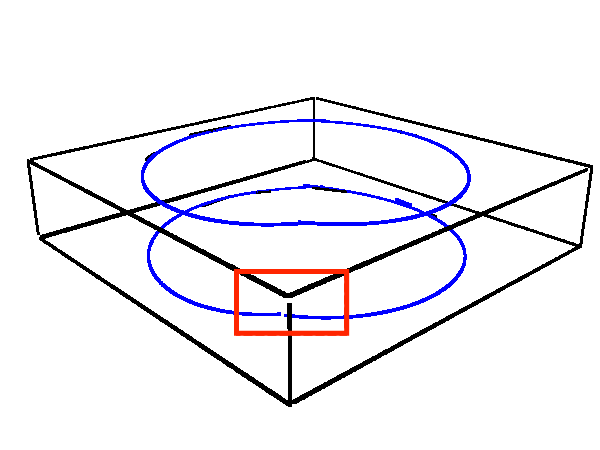}
        \label{fig:ablation_e}
    }
    \subfloat[GT Edge]{%
        \includegraphics[width=0.3\linewidth]{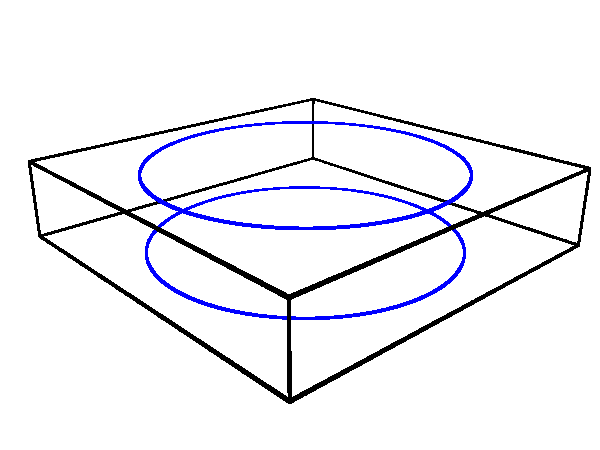}
        \label{fig:ablation_cad}
    }
    \vspace{-0.5em}
    \caption{\textbf{Qualitative ablation on different component of our parametric edge extraction.} The absence of any module in our edge extraction process results in incomplete or noisy qualitative outcomes.}
    \label{fig:ablation_ABC}
    \vspace{-1.5em}
\end{figure}

Furthermore, we also conduct ablation studies in~\tabref{tab:ablation_extraction} and~\figref{fig:ablation_ABC} to evaluate the impact of different components in our edge extraction algorithm. 
These experiments were performed on the ABC-NEF dataset using the DexiNed detector.
First, the removal of the query point shifting step leads to a significant drop in both recall and precision. 
This indicates that our point-shifting step significantly refines the query points locations. 
Second, excluding Bézier curves results in a decline in completeness (\figref{fig:ablation_ABC}~(c)), 
showing that curves are necessary for edge reconstruction. 
Third, omitting the edge merging step leads to redundant small line segments, as evident in~\figref{fig:ablation_ABC}~(d). 
Finally, the removal of endpoint merging impairs connectivity between edges, as shown in ~\figref{fig:ablation_ABC}~(e). 

\begin{table}[h]
\centering
\scalebox{0.83}{

\begin{tabular}{c|l|ccccc}
 & Method & Acc$\downarrow$ & \multicolumn{1}{l}{Comp$\downarrow$} & R5$\uparrow$ & P5$\uparrow$  & F5$\uparrow$ \\ \hline
a & Ours                & \textbf{8.8} & 8.9          & \textbf{56.4}    & 62.9   & \textbf{59.1} \\ \hline
b & w/o point shifting  & 15.3         & 9.9          &  29.2    & 18.7    & 22.2          \\
c & w/o B'ezier curve   &  9.4 & 12.1 & 54.2 & \textbf{65.8} & 59.0       \\
d & w/o edge merging      &  10.3 & \textbf{8.7} & 53.8 & 45.3 & 48.6       \\
e & w/o endpoints merging & 9.3 & 9.0 & 51.5 & 57.7 & 54.0        
\end{tabular}
}
\vspace{-0.5em}
\caption{\textbf{Ablation studies on different component of parametric edge extraction on ABC-NEF~\cite{ye2023nef} with DexiNed~\cite{poma2020dense}.} Our parametric edge extraction approach with all components achieves the optimal balance between accuracy and completeness.}
\label{tab:ablation_extraction}
\vspace{-0.5em}
\end{table}

\begin{figure}[!t]
\centering
\setlength\tabcolsep{2pt} %
\scalebox{0.82}{
\begin{tabular}{cc|cc}
\includegraphics[width=2.3cm]{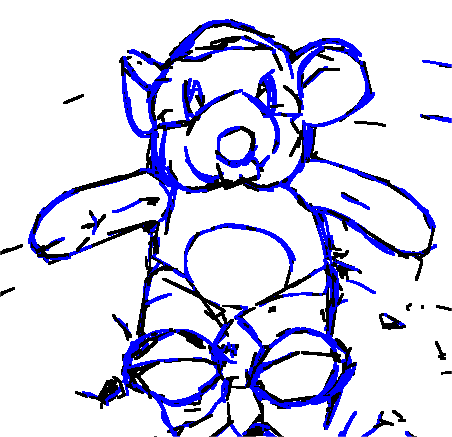} & \includegraphics[width=2.3cm]{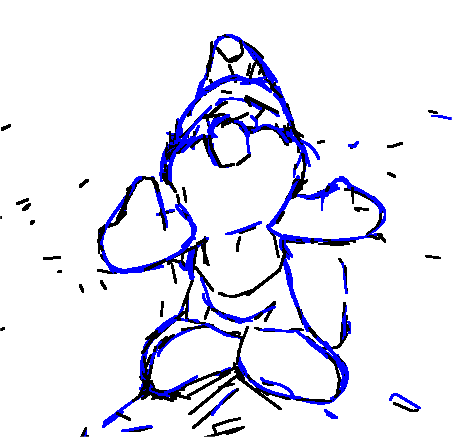}
& \includegraphics[width=2.3cm]{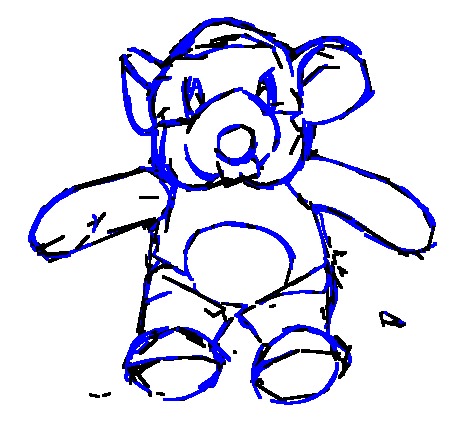} & \includegraphics[width=2.3cm]{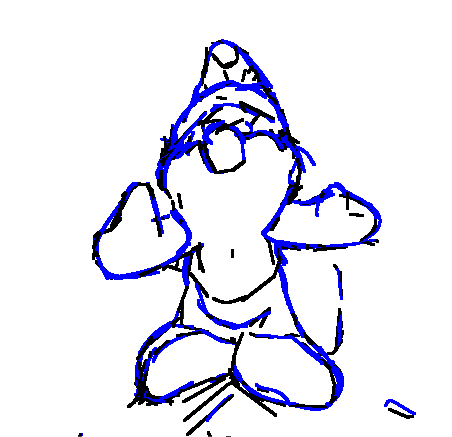}\\
\multicolumn{2}{c}{w/o edge refinement} & \multicolumn{2}{c}{w/ edge refinement}\\
\end{tabular}
}
\vspace{-0.5em}
\caption{\textbf{Ablation study on edge refinement.} Our edge refinement effectively eliminates the majority of noisy edges in background areas.
}
\label{fig:edge_refine}
\end{figure}

\boldparagraph{Edge Refinement}
\label{sec:edge_refine}
In~\figref{fig:edge_refine}, we study the effectiveness of our edge refinement module. When input edge maps contain some noises in dark scenes, 
our initial 3D edge map, without the edge refinement, exhibits some artifacts. However, the edge refinement module markedly mitigates artifacts, achieving a balance between recall and precision.

\begin{figure}[!t]
\centering
\vspace{-0.5em}
\setlength\tabcolsep{1.0pt} %
\footnotesize
\scalebox{0.87}{
\begin{tabular}{cccc}
2D Image & MonoSDF~\cite{yu2022monosdf} & MonoSDF w/ Ours & GT Mesh\\
\includegraphics[width=2.3cm]{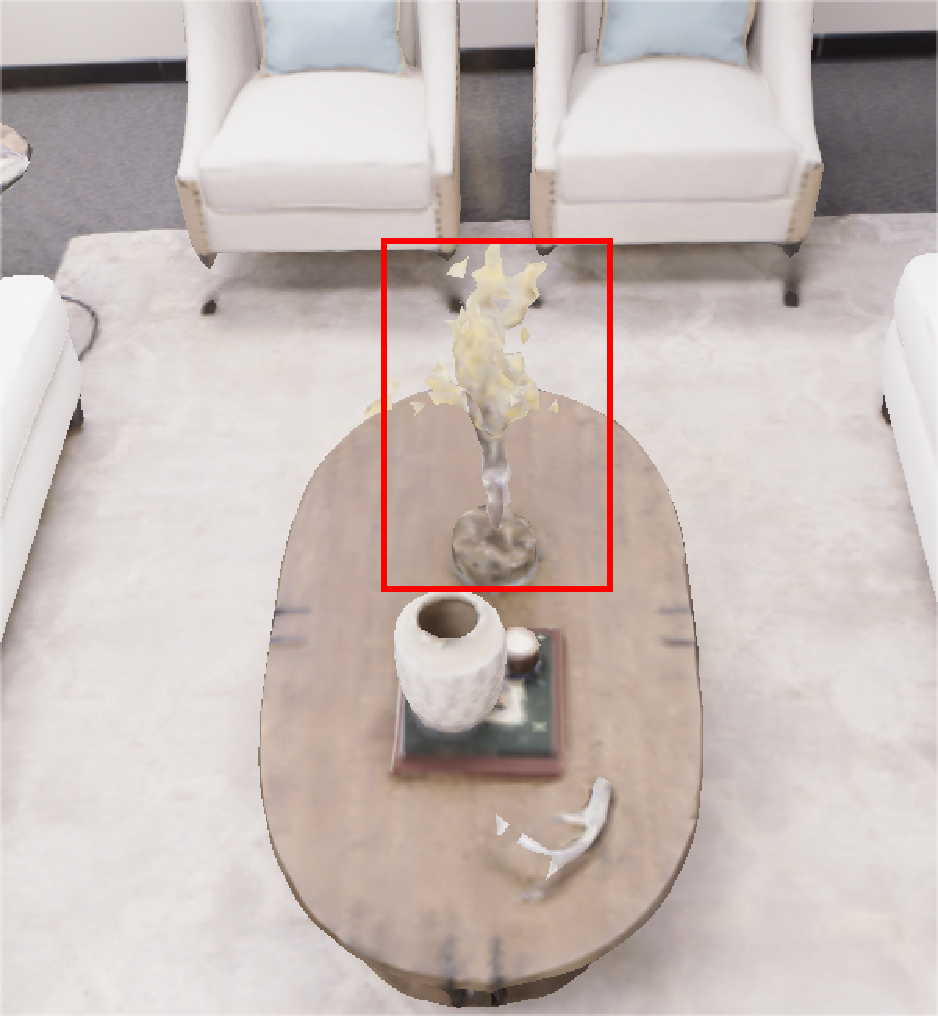} & \includegraphics[width=2.3cm]{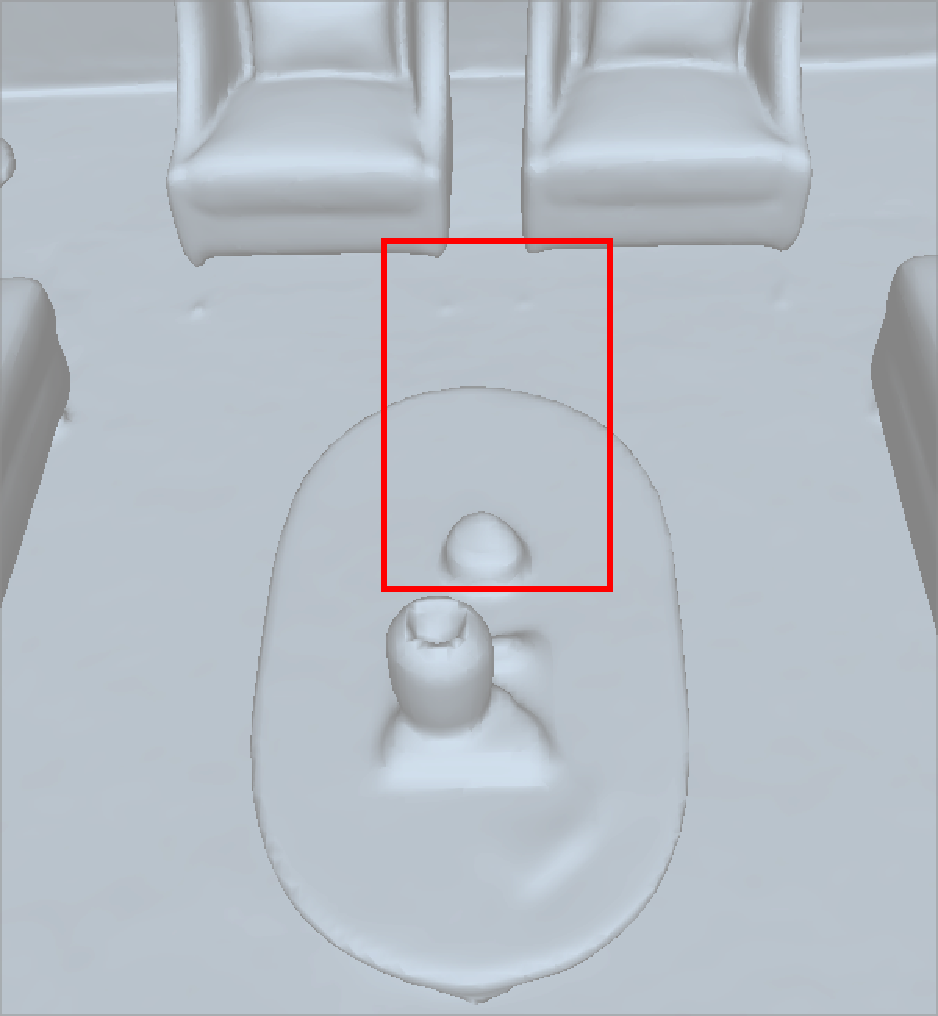} & \includegraphics[width=2.3cm]{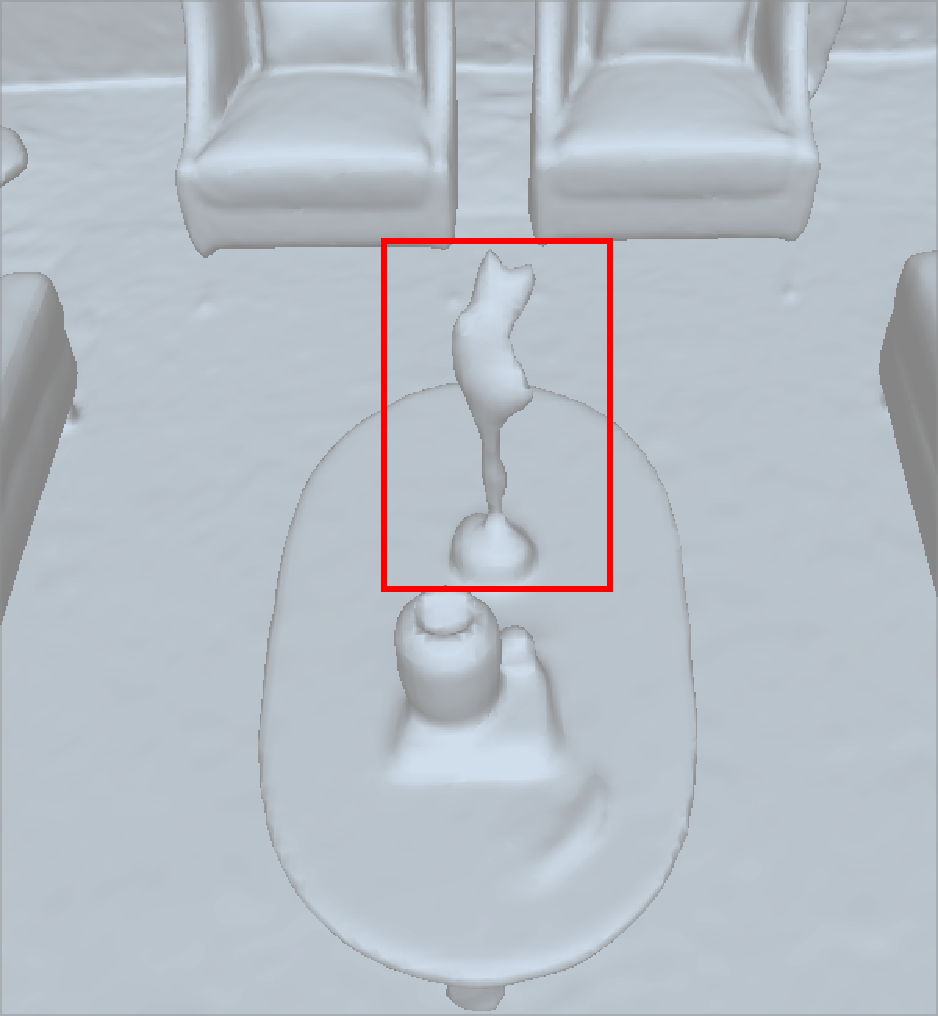} & \includegraphics[width=2.3cm]{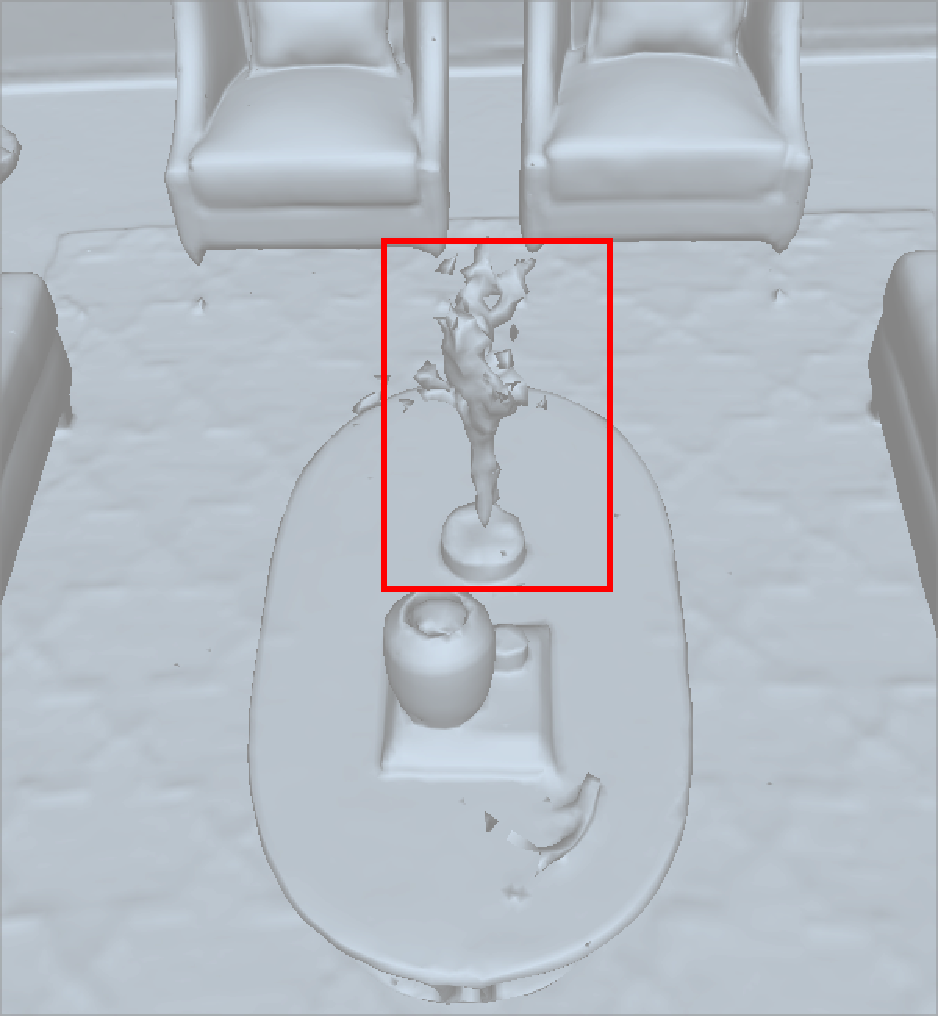}\\
\includegraphics[width=2.3cm]{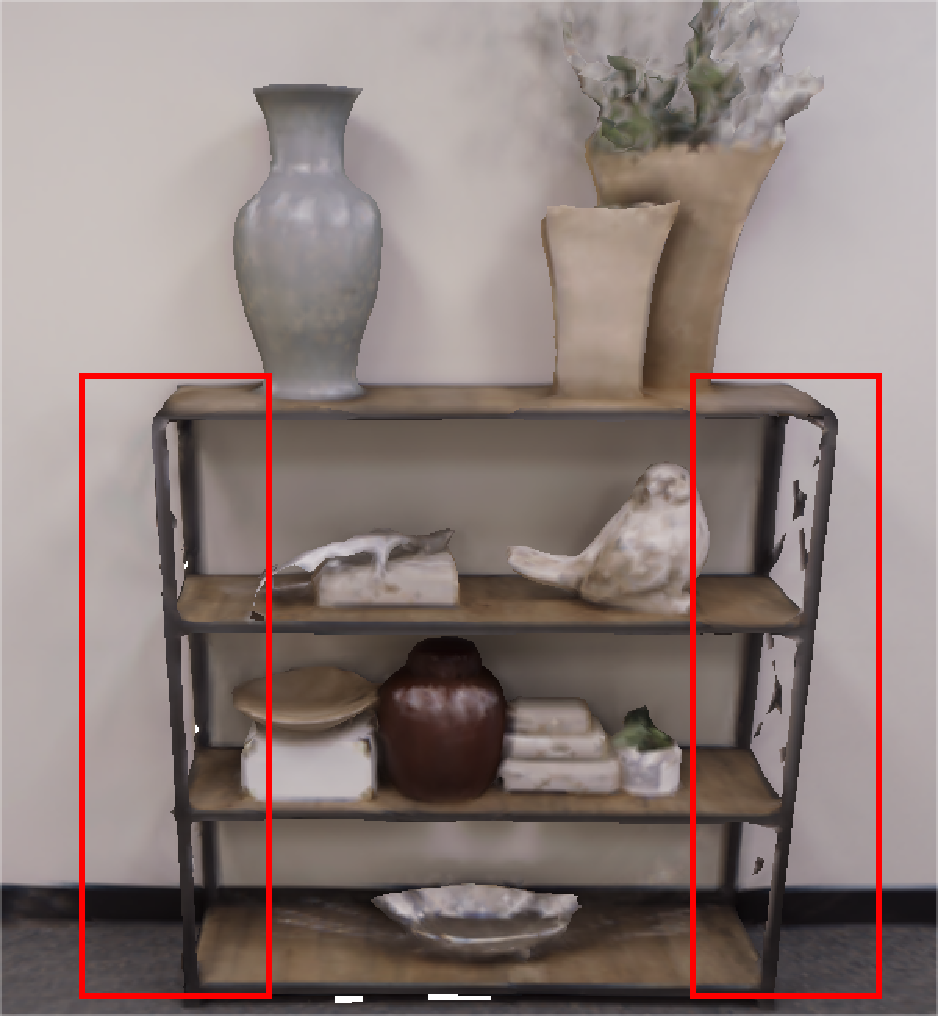} & \includegraphics[width=2.3cm]{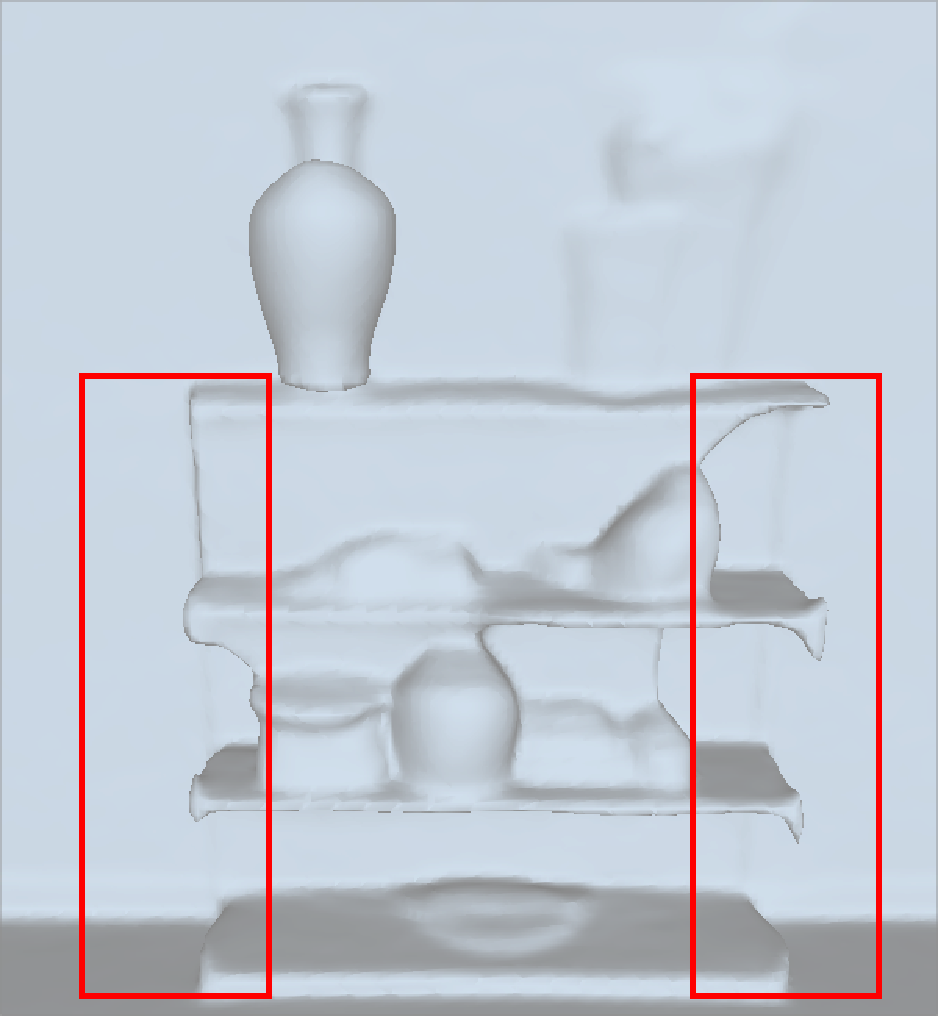} & \includegraphics[width=2.3cm]{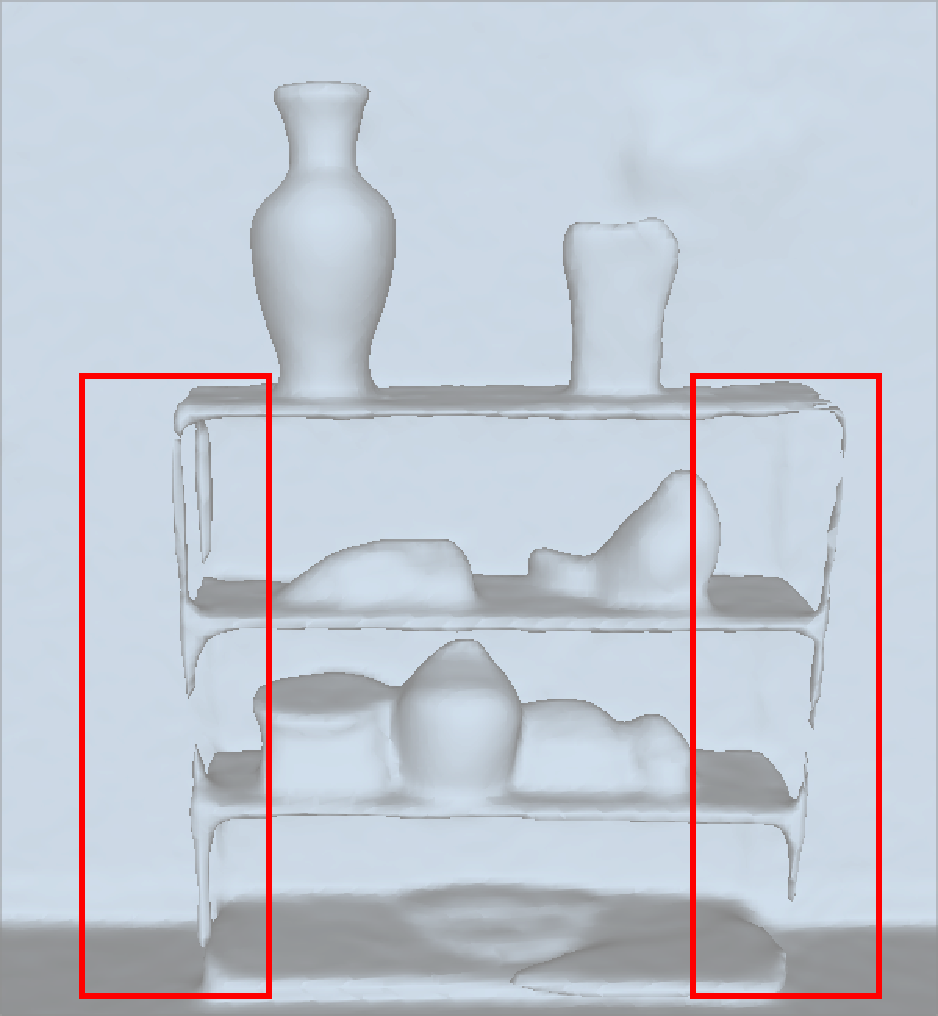} & \includegraphics[width=2.3cm]{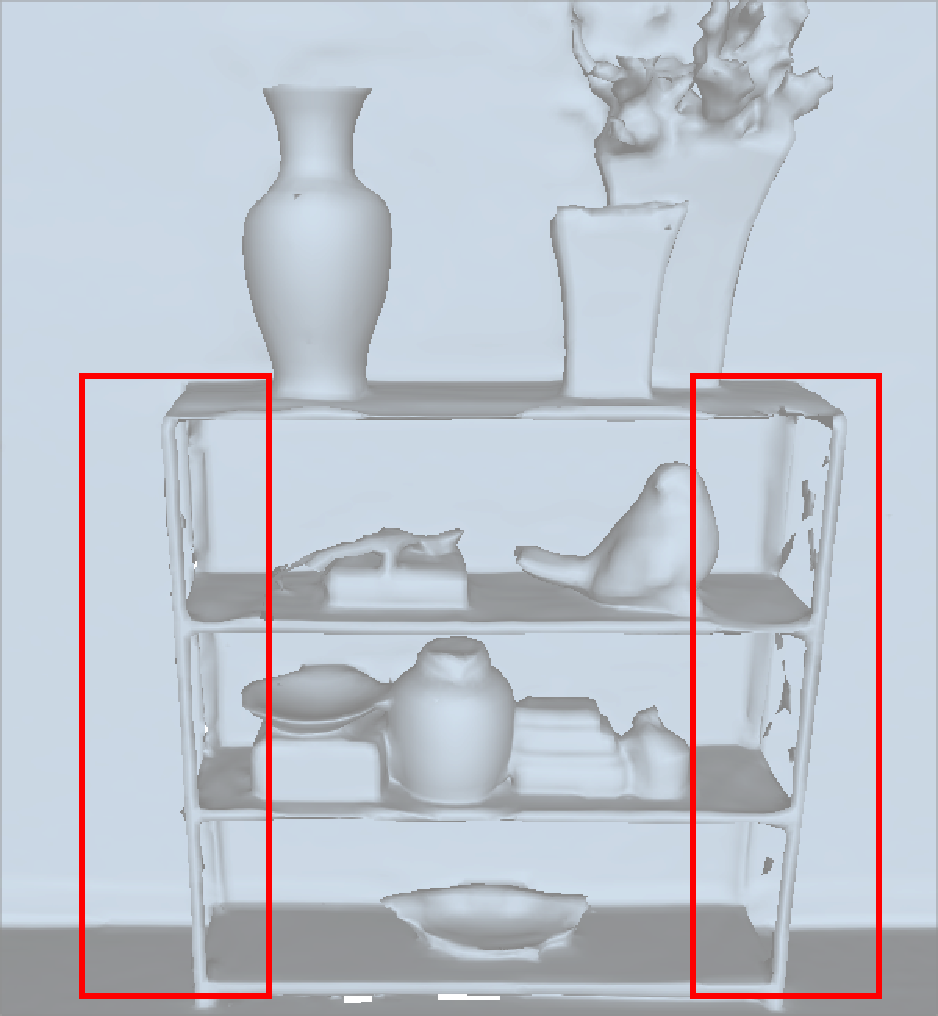}\\
\end{tabular}
}
\vspace{-0.5em}
\caption{\textbf{Dense surface reconstruction on Replica~\cite{straub2019replica}.} Utilizing our trained UDF MLP for initialization enables MonoSDF to capture more geometric details, such as the vase in the top row, the shelf in the bottom row.
}
\label{fig:replica_dense}
\vspace{-1.5em}
\end{figure}

\subsection{Application on Dense Surface Reconstruction}
\label{sec:dense_reconstruction}
Our method has demonstrated its proficiency in reconstructing 3D edges across a diverse range of scenarios. Building on this success, we further explore the potential of our learned representation to benefit other tasks. 
A particularly relevant area is dense surface reconstruction.

As shown in~\figref{fig:replica_dense}, the recent neural-implicit surface reconstruction approach MonoSDF~\cite{yu2022monosdf} can show decent reconstruction results from only posed multi-view images.
However, we notice that they still struggle to capture detailed geometry.
To address this, we integrate our method into the MonoSDF pipeline. 
Specifically, we initialize the geometry MLPs of MonoSDF with our pre-trained UDF MLPs. We can clearly see that such a simple integration can enhance the recovery of geometric details.

\section{Conclusions}

We introduced EMAP, a 3D neural edge reconstruction pipeline that learns accurate 3D edge point locations and directions implicitly from multi-view edge maps through UDF and abstracts 3D parametric edges from the learned UDF field. Through extensive evaluations, EMAP demonstrates remarkable capabilities in CAD modeling and in capturing detailed geometry of objects and scenes. Furthermore, we show that our learned UDF field enriches the geometric details for neural surface reconstruction.

{
    \small
    \bibliographystyle{ieeenat_fullname}
    \bibliography{venue,main}
}

\maketitlesupplementary
\renewcommand{\thesection}{\Alph{section}}
\setcounter{section}{0}

\section{More Method Details}
\subsection{Unbiased Density-based Edge Neural Rendering} 

The detailed derivation of $\hat{E}$ along the camera ray $\mathbf{r}$ in the main paper is given as
\begin{equation}
\begin{aligned}
\hat{E}(\mathbf{r}) & = \int_0^{+\infty} w(t) d t \\
& =\int_0^{+\infty} T(t)\cdot\sigma_u(t) d t \\
& =\int_0^{+\infty} {\exp\left(-\int_{0}^{+\infty} \sigma_u(v) d v\right)} \cdot \sigma_u(t) d t \\
& =\int_0^{+\infty} \frac{d}{d v}\left[-\exp \left(-\int_0^t \sigma_u(v) d v\right)\right] d t \\
& =1-\exp \left(-\int_{0}^{+\infty} \sigma_u(v) d v\right) \\
& =1-T(+\infty) \,.
\end{aligned}
\label{eq:edge}
\end{equation}

Note that this equation occurs in the continuous domain. To obtain the discrete counterparts of the edge rendering function, we adopt the same discrete accumulated transmittance approach used in NeRF~\cite{mildenhall2020nerf}.
\begin{equation}
    \hat{E}(\mathbf{r}) = 1 - T_N,\;\;\; T_N=\exp \left(-\sum_{j=1}^{N-1} {\sigma_u}_j \cdot \left( t_{j+1} - t_{j}\right)\right)\,,
\end{equation}
where $N$ is the number of samples on the ray, $T_N$ denotes the accumulated transmittance at the furthest sampling point from the camera center, and $t_j$ is the $j$-th sample on the camera ray $\mathbf{r}$. 

\begin{figure}[!t]
  \begin{center}
  \includegraphics[width=1.0\linewidth]{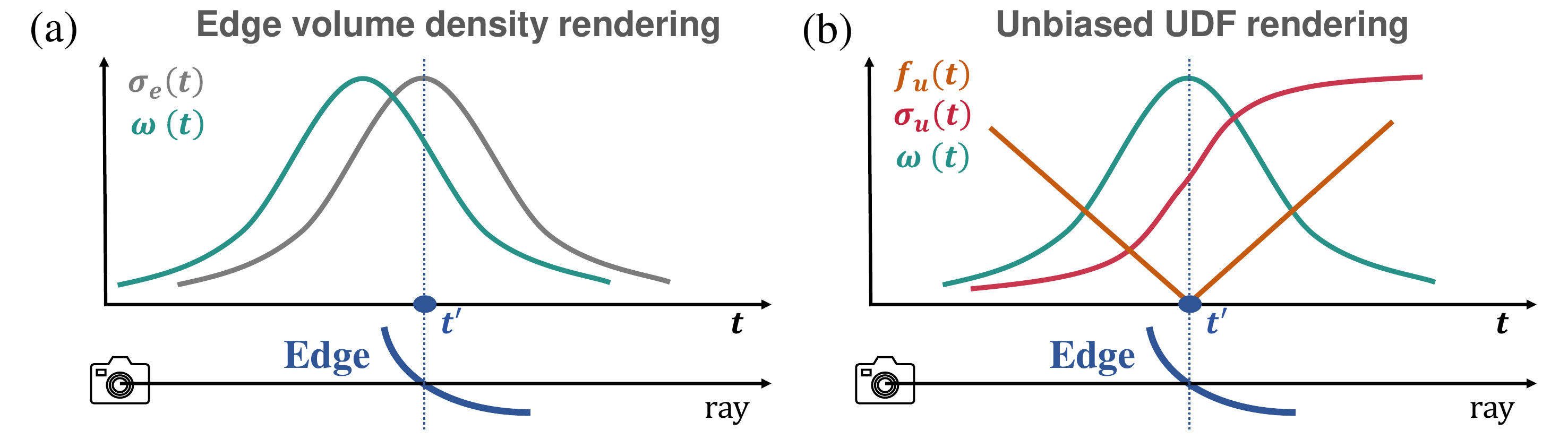}  
  \end{center}
  \vspace{-0.5em}
  \caption{\textbf{Illustration of edge volume density rendering (left) and unbiased UDF rendering (right).} In edge volume density rendering, the weight function does not peak at the intersection point $t'$. Conversely, in unbiased UDF rendering, the weight function's peak aligns precisely with $t'$.}
  \label{fig:biased_unbiased}
\end{figure}

As outlined in the main paper, NEF~\cite{ye2023nef} utilizes edge volume density for representing edges, assigning high-density values $\sigma_e$ for edge points. However, this approach, similar to the naive solution of NeuS~\cite{wang2021neus}, does not align the local maximum of the weight function to the actual intersection point of the camera ray with the edges (\figref{fig:biased_unbiased} (a)), leading to inaccuracies in 3D edge reconstruction.
To address this issue, we incorporate unbiased UDF rendering~\cite{long2023neuraludf} into our density-based edge rendering framework, as depicted in \figref{fig:biased_unbiased} (b). As mentioned in the main paper, the monotonically increasing density function $\sigma_u$~\cite{long2023neuraludf} is formulated as
\begin{equation}
\sigma_u(t) =\Psi(t)\cdot\Omega_s\left(f_u(\mathbf{r}(t))\right)+(1-\Psi(t)) \cdot \Omega_s\left(-f_u(\mathbf{r}(t))\right) \,,
\end{equation}
where $\Psi(t)$ is a differentiable visibility function~\cite{long2023neuraludf} that identifies the intersection point of the camera ray with the edges, and $\Omega_s$ is a monotonic density function that is introduced in NeuS~\cite{wang2021neus} using the Sigmoid function $\Phi_k(x)=\left(1+e^{-k x}\right)^{-1}$. The functions $\Psi(t)$ and $\Omega_s$ are formulated as
\begin{equation}
\Psi\left(t_j\right) =\prod_{i=1}^{j-1}\left(1-h\left(t_i\right) \cdot m\left(t_i\right)\right),
\end{equation}
\begin{equation}
m\left(t_j\right) = \begin{cases}0, \quad \cos \left(\theta_{j+1}\right)<0 \\
1, \quad \cos \left(\theta_{j+1}\right) \geq 0\end{cases},
\end{equation}
\begin{equation}
   \Omega_s = \max \left(\frac{-\frac{\mathrm{d} \Phi_k}{\mathrm{~d} t}(f_u(\mathbf{r}(t)))}{\Phi_k(f_u(\mathbf{r}(t)))}, 0\right),
\end{equation}
where $m\left(t_j\right)$ is a masking function that masks out the points behind the intersection point, with $\theta_{j+1}$ being the angle between the ray and the gradient of $f_u$ at $\mathbf{r}(t+1)$. The probability of intersection $h(t_j)$ is 1 when the ray at $t_j$ intersects with edges and is given by a logistic distribution $\phi_\beta(x)$,
\begin{equation}
    h\left(t_j\right)=1-\exp \left(-\alpha \cdot \phi_\beta\left(f_u\left(\mathbf{r}\left(t_j\right)\right)\right) \cdot \delta_j\right),
\end{equation}
\begin{equation}
    \phi_\beta(x)=\beta e^{-\beta x} /\left(1+e^{-\beta x}\right)^2,
\end{equation}
where $\alpha = 20$ and $\beta$ is a learned parameter.

\begin{algorithm}[!ht]
    \caption{\textbf{Point Connection}}
    \label{alg:point_connection}
    \begin{algorithmic}[1]
        \Require Set of potential edge points $\mathcal{U}$ with position and local edge direction, search distance threshold $d_t$, direction similarity threshold $s_t$, and NMS ratio $n_r$.
        \Ensure Collection of edge point groups.
        
        \While{$\mathcal{U} \neq \emptyset$}
            \State Select a point $p$ randomly from $\mathcal{U}$ and remove $p$ from $\mathcal{U}$.
            \State Initialize a new edge point group $G$ with $p$.
            \For{each direction case: forward ($+$) and backward ($-$)}
                \State Initialize an empty list $S$ for storing similarities.
                \ForAll{points $\hat{p}$ in $\mathcal{U}$ within distance $d_t$ of $p$}
                    \State Calculate similarity $s_{\hat{p}} \leftarrow \cos(\vec{p}, \vec{\hat{p}}) \cdot \text{sign}(direction)$.
                    \State Append $s_{\hat{p}}$ to $S$.
                \EndFor
                \State Find point $\hat{p}_i$ with the highest similarity in $S$.
                \If{$s_{\hat{p}_i} > s_t$}
                    \State Update $p$ to $\hat{p}_i$, remove $\hat{p}_i$ from $\mathcal{U}$.
                    \State Add $\hat{p}_i$ to group $G$.
                    \State Apply NMS: Remove points from $\mathcal{U}$ with similarity greater than $n_r \cdot s_{\hat{p}_i}$.
                \EndIf
            \EndFor
        \EndWhile
        \State \Return the set of all edge point groups formed.
    \end{algorithmic}
\end{algorithm}

\subsection{Point Connection Algorithm}
In the point initialization step, we set the UDF threshold to $\epsilon'$ and normalize the 3D voxel grid to a range of [-1, 1], with a grid size of $M^3$. In the point shifting step, we perform $T$ iterations. The point connection algorithm is detailed in Alg.~\ref{alg:point_connection}.

In this algorithm, we establish thresholds for search distance ($d_t$), direction similarity ($s_t$), and Non-Maximum Suppression (NMS) ratio ($n_r$). The process begins with initializing the set of unvisited points ($\mathcal{U}$) with all potential edge points. A random target point $p$ is then selected from $\mathcal{U}$, removed from the set, and added to an edge point group. We identify adjacent points $\hat{p}$ from $\mathcal{U}$ within the search distance $d_t$ from $p$. The edge point group, being undirected, is extended in two subgroups: one extending forward along the edge direction, and the other extending backward along the inverse edge direction. For each adjacent point, we compute its similarity ($s$) with the target point as follows
\begin{equation}
s_j = \begin{cases} \cos(\overrightarrow{pp_{j}}, l(p)), \quad \text{forward} \\
-\cos(\overrightarrow{pp_{j}}, l(p)), \quad \text{backward} \end{cases},
\end{equation}
where $j$ indexes the adjacent points. The similarity $s_j$ is calculated using the edge direction for the forward case and the inverse edge direction for the backward case. The adjacent point with the highest similarity ($s_i$) is selected as the candidate. If $s_i$ exceeds $s_t$, the candidate point is updated as the new target point, removed from $\mathcal{U}$, and added to the edge point group. Redundant points with similarity greater than $n_r \cdot s_i$ are also removed from $\mathcal{U}$. If $s_i$ is not greater than $s_t$, the growth of the current point connection halts, and a new target point is selected from $\mathcal{U}$ to initiate a new edge point group. This process is repeated for each edge point group, progressively extending it until no further target points can be added. Finally, we obtain connected edge points from each edge point group.

\subsection{Edge Fitting}
\begin{figure}[!t]
  \begin{center}
  \includegraphics[width=1.0\linewidth]{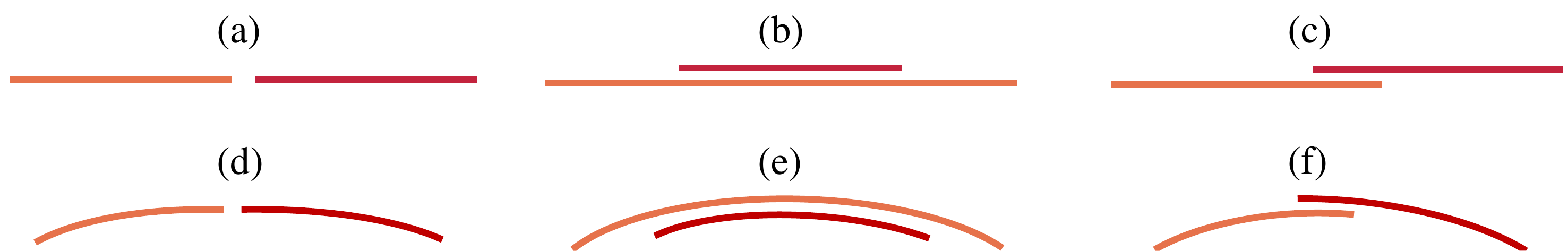}  
  \end{center}
  \vspace{-0.5em}
  \caption{\textbf{Illustration of edge merging.} (a)-(c) demonstrate line segment merging, and (d)-(f) show curve merging. In all cases, the candidate edges are closely aligned in terms of shortest distance, with similar line directions or curvatures at their nearest points.}
  \label{fig:edge_merging}
\end{figure}

As introduced in the main paper, the merging of line segments and Bézier curves is based on two primary criteria: the shortest distance ($d_s$) between candidate edges and the curvature similarity ($s_c$) at their nearest points. \figref{fig:edge_merging} illustrates distinct cases for both line segments and curves. These criteria ensure that merging occurs only between closely situated edges with similar features.

\section{More Experiment Details} 
\subsection{Implementation Details of UDF Field}
We utilize one MLP with an absolute activation function in the last layer to learn UDF values. Following previous works~\cite{wang2021neus, yariv2021volume}, we include a skip connection that links the output of the fourth layer with its input. Positional encoding is applied using 10 frequencies. Following NeuralUDF~\cite{long2023neuraludf}, we initially sample 64 points uniformly and then perform iterative importance sampling five times to refine the sampling based on the UDF values. In each importance sampling iteration, 16 points are sampled. Additionally, we normalize all scenes into a unit box. This normalization ensures that the hyperparameters for 3D parametric edge extraction are generally invariant to the scale of the scenes. $\lambda$ in the loss function is set to $0.1$ for ABC-NEF dataset, and $0.01$ for other dataset.

\subsection{Implementation Details of 3D Parametric Edge Extraction}
Similar to LIMAP~\cite{liu20233d}, our parametric edge extraction approach involves several hyperparameters within each module. 
However, as we normalize all scenes into a unit box within our UDF field, our default settings typically suffice.
The UDF threshold ($\sigma'$) and voxel grid size ($M$) are adjusted according to scene complexity. For example, we use $\sigma' = 0.02$ and $M = 128$ for the synthetic ABC-NEF~\cite{ye2023nef} dataset, while for other datasets like DTU~\cite{aanaes2016large}, Replica~\cite{straub2019replica}, and Tanks \& Temples~\cite{knapitsch2017tanks}, we opt for $\sigma' = 0.01$ and $M = 256$. 
We set the point shifting iteration ($T$) to 2, which yields optimal performance, with further details in \secref{sec:iter}. 
In the edge direction extraction step, the shift set ${\delta}_N$ is randomly sampled from the range $\left[-5\times 10^{-3}, 5\times 10^{-3}\right]$ with 50 samples. In point connection, the parameters $d_t$, $s_t$, and $n_r$ are set to $\frac{10}{M}$, $0.97$, and $0.95$, respectively. For edge fitting, we require a minimum of 5 inlier points for robust line segment fitting and at least 4 for Bézier curve fitting. During edge merging, we apply criteria of $d_s = \frac{5}{M}$ and $s_c = 0.98$. Endpoints within a distance of $d_e=\frac{2}{M}$ are merged into shared endpoints. Edge refinement is applied specifically to real-world datasets to mitigate the effect of noise in the input edge maps.

\subsection{Ground-truth Edge Point Generation on DTU}
As mentioned in the main paper, the DTU dataset provides dense ground-truth point clouds that can be further processed into edge points. To meet multi-view consistent edge requirements, we select scan 37, scan 83, scan 105, scan 110, scan 118, and scan 122 as our evaluation scans. Inspired by \cite{bignoli2018edgegraph3d} that builds edge point matching using the sparse 3D points from Structure-from-Motion, we generate edge points by projecting the ground-truth dense points onto images and then cross-comparing these projections with observations on 2D edge maps. This process allows us to filter out points that do not correspond to edges. To ensure accuracy in the ground-truth edge points, we manually set thresholds for each scan and meticulously remove any floating points.

\section{More Experiment Results} 

\subsection{Edge Representations} 
\begin{figure}[!t]
    \centering
    \subfloat[CAD Model]{%
        \includegraphics[width=0.3\linewidth]{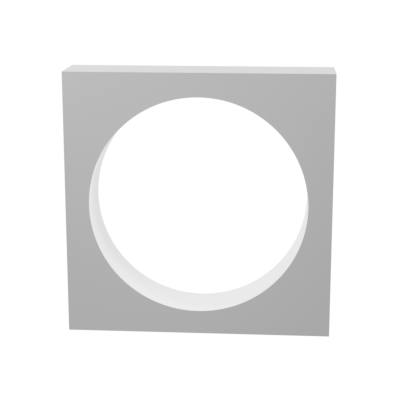}
    }
    \subfloat[Edge Map]{%
        \includegraphics[width=0.3\linewidth]{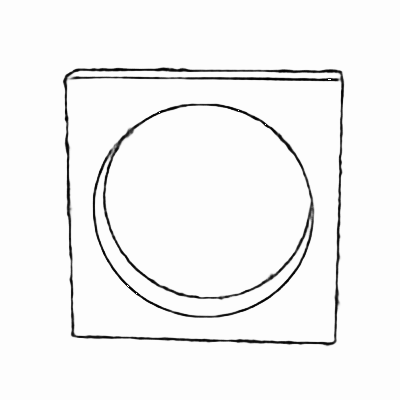}
    }
    \subfloat[SDF Edge]{%
        \includegraphics[width=0.3\linewidth]{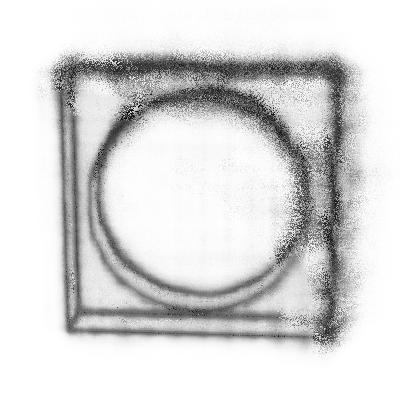}
    } \\
    \subfloat[UDF Edge]{%
        \includegraphics[width=0.3\linewidth]{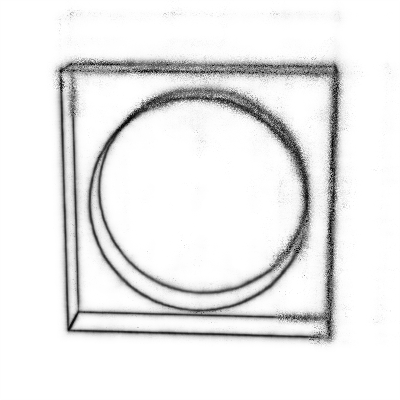}
    }
    \subfloat[SDF Normal]{%
        \includegraphics[width=0.3\linewidth]{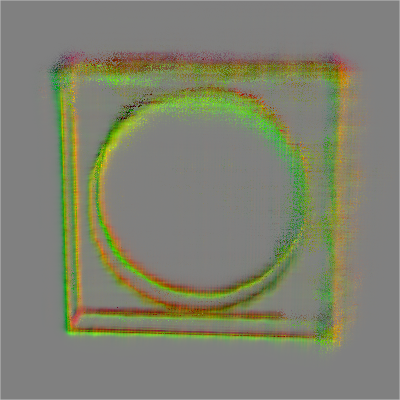}
    }
    \subfloat[UDF Normal]{%
        \includegraphics[width=0.3\linewidth]{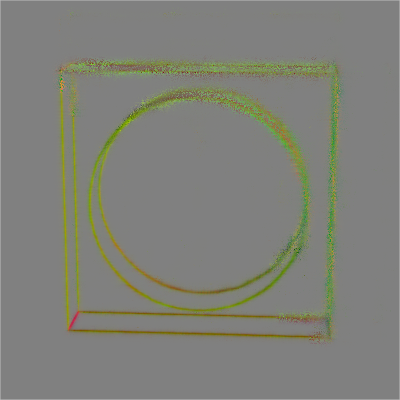}
    }
    \vspace{-0.5em}
    \caption{\textbf{Ablation study on SDF and UDF.} Representing edges using SDF leads to blurred edge rendering, as edges do not have a clear definition of inside and outside. In contrast, UDF is well suited for modeling edge distance field.
    }
    \label{fig:sdf_udf}
\end{figure}

We evaluate the effectiveness of SDF and UDF for edge representation.
\begin{table}[!t]
\centering
\scalebox{0.9}{

\begin{tabular}{c|c|cccccc}
  & Ratio & Acc$\downarrow$ & Comp$\downarrow$ & Norm $\uparrow$ & R5$\uparrow$ & P5$\uparrow$  & F5$\uparrow$ \\ \hline
a & 25\% & 7.9 & 8.9 & 95.3 & \textbf{57.1} & \textbf{64.7} & \textbf{60.3}\\
b & 50\%           & 8.8     & \textbf{8.9} & \textbf{95.4}  & 56.4 & 62.9  &  59.1 \\
c & 75\%           & \textbf{7.7}     & 10.3 & 94.8  & 53.2 & 58.3  &  55.3 \\
\end{tabular}
}
\vspace{-0.5em}
\caption{Ablation studies on ray sampling strategy on ABC-NEF. Performance metrics for 25\% sampling (a) and 75\% sampling (c) are averaged over \textit{valid scans} only, as 25\% sampling results in invalid outcomes for scan 9685, and 75\% sampling is ineffective for scans 2412 and scan 5109.}
\label{tab:ablation_sampling}
\end{table}

As illustrated in~\figref{fig:sdf_udf}, the edge map and normal map rendered using SDF tend to be noisy and lack precision. This issue primarily arises because SDF rendering requires a clear definition of inside and outside, which edges do not inherently possess. In contrast, UDF excels at representing both closed and open surfaces in volume rendering, making them more suitable for modeling the edge distance field. Consequently, UDF-rendered edge and normal maps exhibit greater precision. In comparison, representations based on edge volume density, like those in NEF~\cite{ye2023nef} that often rely on fitting-based extraction methods, are more prone to inaccuracies.
Therefore, we select UDF as our edge representation.  

\subsection{Ray Sampling Strategy} 
To evaluate the effectiveness of our ray sampling strategy, we conduct ablation studies on the sampling ratio in \tabref{tab:ablation_sampling}. The sampling ratio in edge regions is then incrementally increased from 25\% to 75\%. The results indicate that a 50\% sampling ratio is the only strategy that consistently yields complete results across all scans, while 25\% and 75\% sampling ratios fail to reconstruct all scans.
Besides, the commonly used uniform sampling strategy is largely ineffective, predominantly due to occlusion-related issues.

\begin{table}[!t]
\centering
\scalebox{0.9}{

\begin{tabular}{c|c|cccccc}
  & Iter. & Acc$\downarrow$  & Comp$\downarrow$ & Norm$\uparrow$ & R5$\uparrow$ & P5$\uparrow$  & F5$\uparrow$ \\ \hline
a & 0     & 14.4            & 9.4      & 94.3 &   32.7 & 26.2 & 28.3        \\

c & 1     & 9.9            & 9.0     & 95.1   &    49.6 & 53.8 & 51.2     \\

d & 2     & \textbf{8.8}             & \textbf{8.9}   & \textbf{95.4} & \textbf{56.4} & \textbf{62.9} & \textbf{59.1}   \\

e & 3     & 9.1             & 9.1    &95.4   &  53.5 & 59.8 & 56.1          
\end{tabular}
}
\vspace{-0.5em}
\caption{Ablation studies on point shifting iterations on ABC-NEF.}
\label{tab:point_shifting_iter}
\end{table}

\subsection{Point Shifting Iteration}\label{sec:iter} 
We conduct analytical experiments to assess the impact of point-shifting iterations, as presented in \tabref{tab:point_shifting_iter}. We empirically find that both accuracy and completeness are maximized when the iteration count, $T$, is set to 2. As discussed in the main paper, the absence of point shifting ($T = 0$) results in significantly lower accuracy due to the prediction of redundant edge points. Conversely, increasing the iteration count ($T = 3$) leads to a degradation in performance, attributable to the introduction of noise from over-iteration.

\begin{table}[!t]
\centering

\begin{tabular}{c|ccc|ccc}
\multirow{2}{*}{Res} & \multicolumn{3}{c|}{w/o point shifting} & \multicolumn{3}{c}{w/ point shifting} \\
& Acc$\downarrow$ & Comp$\downarrow$ & F5$\uparrow$ & Acc$\downarrow$ & Comp$\downarrow$ & F5$\uparrow$ \\
\hline

64 & 15.1 & 31.3 & 13.8 & 8.8 & 29.0 & 37.8\\

128 & 14.4 & 9.4 & 28.3 & \textbf{8.8} & 8.9 & \textbf{59.1} \\

256 & 16.5 & \textbf{8.0} & 24.6 & 10.5 & 8.2 & 55.2 \\
\end{tabular}
\vspace{-0.5em}
\caption{Ablation studies on point-shifting with various grid resolutions on ABC-NEF.}
\label{tab:point_res_ablation}

\end{table}

\subsection{Point Shifting vs. Grid Resolution.} 
We perform ablation studies to analyze the effectiveness of point-shifting step with various voxel grid resolutions. \tabref{tab:point_res_ablation} shows that point shifting consistently boosts the performance across various resolutions. Moreover, increasing grid resolution does improve completeness, but could harm accuracy. Additionally, the cubic growth in query points with higher resolutions leads to a substantial computation increase.

\begin{figure}[!t]
  \begin{center}
  \includegraphics[width=1.0\linewidth]{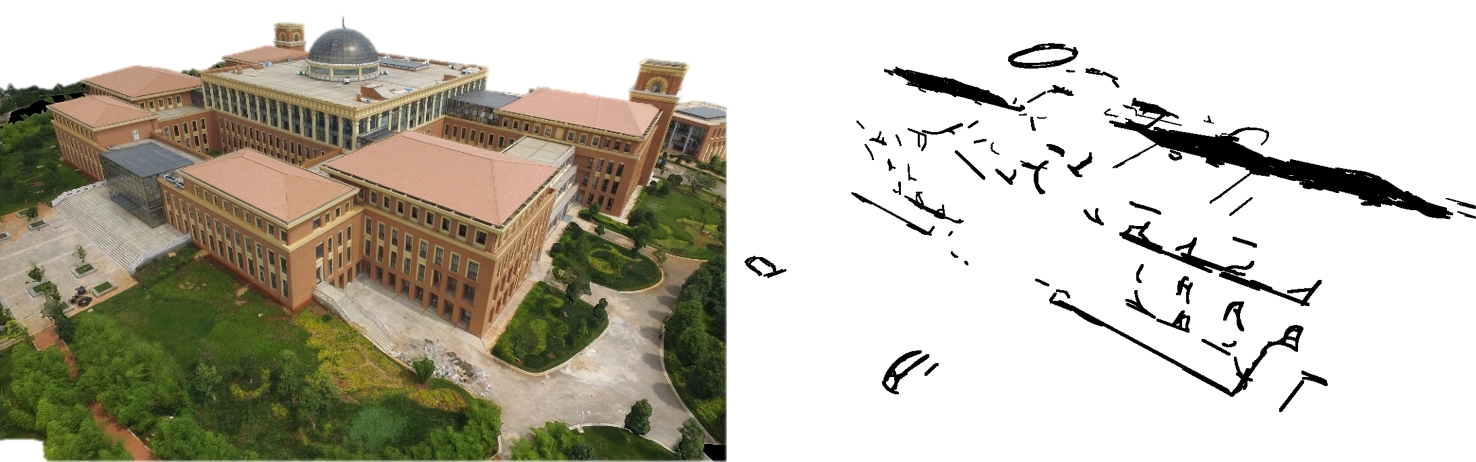}  
  \end{center}
  \vspace{-0.5em}
  \caption{\textbf{A failure case in the large-scale scene.} \textit{left}: RGB image, \textit{right}: reconstructed 3D edges.}
  \label{fig:failure_case}
\end{figure}

\section{Limitations and Future Work}
Our approach successfully achieves comprehensive edge reconstruction across various datasets. However, its application to large-scale datasets is somewhat restricted due to our reliance on edge maps without texture features and the use of vanilla NeRF~\cite{mildenhall2020nerf} MLP, leading to ill-posed reconstruction scenarios (\figref{fig:failure_case}). Incorporating texture information or monocular depth maps~\cite{yu2022monosdf}, and applying more powerful volume rendering methods~\cite{li2022nerfacc, muller2022instant, li2023neuralangelo} can potentially enhance our method's reconstruction capabilities on large-scale datasets.

Additionally, our method is limited to scenes with only view-consistent edges. To handle view-inconsistent edges, one future work is to apply semantic edge detector~\cite{Pu_2021ICCV_RINDNet} to detect those view-inconsistent edges and remove them from 2D edge maps. Alternatively, in our UDF training phase, we could also add a regularization loss to enforce that 3D edges maintain consistency across multiple views.

\end{document}